\PassOptionsToPackage{table,dvipsnames}{xcolor}
\documentclass{muxlab}

\renewcommand*{\backref}[1]{}
\renewcommand*{\backrefalt}[4]{\ifcase#1\relax\or(page #2)\else(pages #2)\fi}

\usepackage{amsmath,amssymb,amsthm}
\usepackage{mathtools}

\crefformat{equation}{(#2#1#3)}
\Crefformat{equation}{(#2#1#3)}
\crefrangeformat{equation}{(#3#1#4)--(#5#2#6)}
\Crefrangeformat{equation}{(#3#1#4)--(#5#2#6)}
\crefmultiformat{equation}{(#2#1#3)}{ and~(#2#1#3)}{, (#2#1#3)}{, and~(#2#1#3)}
\Crefmultiformat{equation}{(#2#1#3)}{ and~(#2#1#3)}{, (#2#1#3)}{, and~(#2#1#3)}

\usepackage{url}
\def\figref#1{Fig.~\ref{#1}}
\def\Figref#1{Fig.~\ref{#1}}
\def\secref#1{Section~\ref{#1}}
\def\1{\bm{1}}
\def\vb{{\bm{b}}}
\def\vg{{\bm{g}}}
\def\vq{{\bm{q}}}
\def\vu{{\bm{u}}}
\def\vz{{\bm{z}}}
\def\mA{{\bm{A}}}
\def\mB{{\bm{B}}}
\def\mC{{\bm{C}}}
\def\mG{{\bm{G}}}
\def\mH{{\bm{H}}}
\def\mI{{\bm{I}}}
\def\mW{{\bm{W}}}
\def\mX{{\bm{X}}}
\def\mY{{\bm{Y}}}
\newcommand{\Ls}{\mathcal{L}}
\newcommand{\R}{\mathbb{R}}
\newcommand{\softmax}{\mathrm{softmax}}
\newcommand{\methodname}{\texttt{reViT}}
\definecolor{refgray}{gray}{0.45}
\newcommand{\gr}[1]{\textcolor{refgray}{#1}}

\hypersetup{
  linkcolor=muxviolet,
  citecolor=muxviolet,
  urlcolor=muxviolet,
  pdftitle={One Block, Multiple Depths: Recurrent Vision Transformers with Depth-Programmed Experts},
  pdfauthor={Adrian Bulat, Yassine Ouali, Georgios Tzimiropoulos}
}

\graphicspath{{figures/}}

\makeatletter
\patchcmd{\mux@maketitle}{\vskip 0.4cm}{\vskip 0.6cm}{}{\PackageError{main}{title spacing patch failed}{}}
\patchcmd{\mux@maketitle}{\vskip 0.25cm}{\vskip 0.45cm}{}{\PackageError{main}{abstract spacing patch failed}{}}
\makeatother

\title{\fontsize{17.5}{22}\selectfont One Block, Multiple Depths: Recurrent Vision Transformers with Depth-Programmed Experts}

\renewcommand{\authorlist}{{\sffamily\color{muxfg}\textbf{Adrian Bulat}\textsuperscript{1,2}\quad
\textbf{Yassine Ouali}\textsuperscript{1}\quad
\textbf{Georgios Tzimiropoulos}\textsuperscript{1,3}}}

\renewcommand{\affiliationlist}{{\small\textsuperscript{1}Samsung AI Cambridge\quad
\textsuperscript{2}Technical University of Iasi\quad
\textsuperscript{3}Queen Mary University of London}}

\abstract{In this work, we show that a single Transformer block, applied recurrently, can match the accuracy of a full-depth vision encoder at comparable inference FLOPs without intermediate feature distillation. \methodname{} restores depth-specific transformations by representing the FFN at each recurrent depth as a convex combination of a small shared expert bank. A continuous normalized-depth coordinate programs this mixture, defining a resampleable trajectory through FFN parameter space. We evaluate this design in two regimes: supervised ImageNet-1k training and distillation from a DINOv2 teacher. Across both regimes, controlled adaptations identify weight-space merging as the strongest tested MoE family at a matching one-FFN budget, ahead of the token-dispatch and output-mixture alternatives. Trained from scratch, \methodname{}-B/16 attains DeiT III accuracy with about 70\% fewer stored parameters. An 8-experts model distilled using only the teacher's output features retains nearly all of its DINOv2 teacher's linear-probe accuracy and transfers across classification, segmentation, and depth prediction. Elastic-depth training allows one checkpoint (trained model) to operate at multiple tested depths by resampling the same normalized coordinate interval. For fixed-depth deployment, the recurrent block can be materialized as a conventional dense graph, removing online routing and merging without changing the one-FFN-per-depth compute but expanding deployment storage.}

\renewcommand{\beginappendix}{\par
\appendix
\crefalias{section}{appendix}
\crefalias{subsection}{appendix}
\section*{\huge\sffamily\bfseries Appendix}
\addcontentsline{toc}{section}{Appendix}
\vskip 0.5cm}

\begin{document}

\maketitle

\section{Introduction}

Standard Vision Transformers (ViTs) process image tokens through a stack of $L$ transformer blocks, each with its own parameters \citep{dosovitskiy2021image}. Although transformers develop distinct behaviors across depth \citep{ghiasi2022vision,jawahar2019does,valeriani2023geometry}, ViTs exhibit greater representational similarity across layers than comparable CNNs \citep{raghu2021vision}. This tension motivates our central question: can a ViT replace its depth-wise stack with a single recurrent block while retaining the depth-dependent computation needed for accuracy?

Residual and highway networks have long motivated an iterative view of depth \citep{greff2017highway}, while Universal Transformers make weight reuse explicit by repeatedly applying a shared block \citep{dehghani2019universal}. More recently, these ideas have been developed more extensively in language \citep{lan2020albert,bae2025relaxed,tan2023sparse,csordas2024moeut,bae2025mixture}. These results establish the promise of recurrent sharing in language, but leave open which expert formulations best preserve accuracy under full-depth sharing in vision.

For vision, mainstream ViTs are still built as fixed-depth stacks. Only recently, \citet{jacobs2026block} find that, across encoder families, the trained ViT blocks organize into a few contiguous computational phases separated by narrow transitions. Their Raptor model exploits this structure by replacing a depth-$L$ stack with $k \ll L$ distinct blocks applied recurrently, recovering most of the original model's performance with $k{=}4$. Raptor nevertheless leaves open the full-sharing case: its published distilled models retain several distinct blocks, align recurrent segments with intermediate teacher features, and still underperform compared to the network they approximate. 

\begin{figure*}[t]
  \centering
  \begin{subfigure}[t]{0.27\textwidth}
    \centering
    \parbox[c][124pt][c]{\linewidth}{\includegraphics[width=\linewidth]{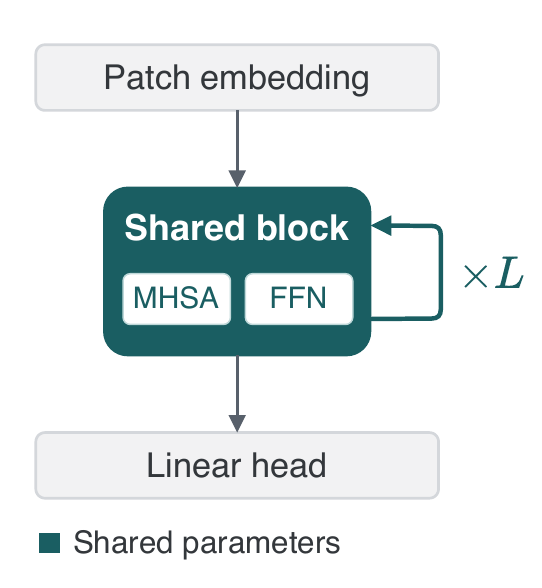}}
      \vspace{-0.2em}
    \caption{Full-depth recurrence.}
    \label{fig:teaser-recurrence}
  \end{subfigure}
  \hfill
  \begin{subfigure}[t]{0.42\textwidth}
    \centering
    \parbox[c][124pt][c]{\linewidth}{\includegraphics[width=\linewidth]{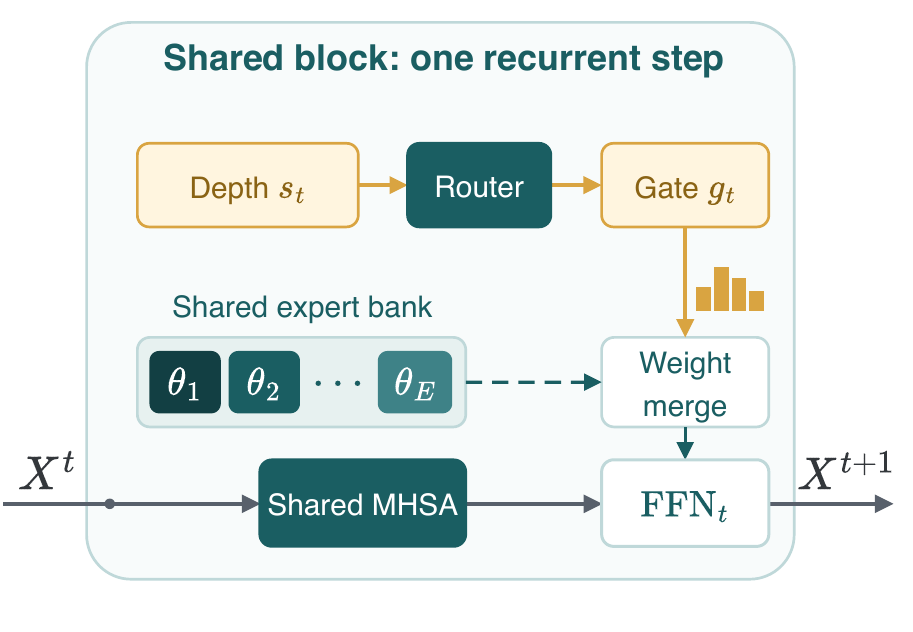}}
      \vspace{-0.2em}
    \caption{Depth-programmed experts.}
    \label{fig:teaser-experts}
  \end{subfigure}
  \hfill
  \begin{subfigure}[t]{0.27\textwidth}
    \centering
    \parbox[c][124pt][c]{\linewidth}{\includegraphics[width=\linewidth]{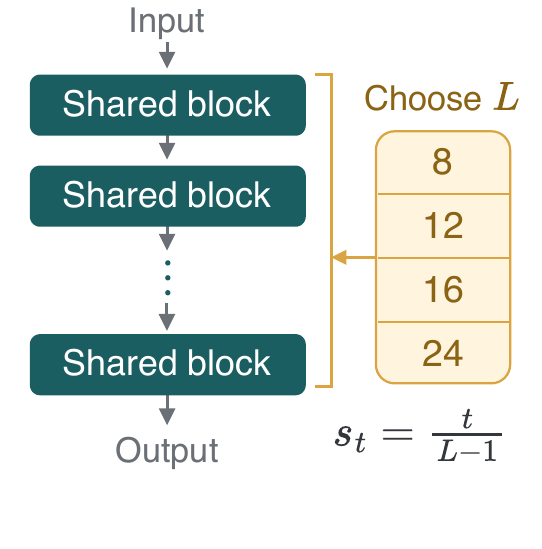}}
      \vspace{-0.2em}
    \caption{Elastic-depth inference.}
    \label{fig:teaser-elastic}
  \end{subfigure}
  \vspace{-0.2em}
  \caption{\textbf{\methodname{} overview.} (a) One shared module runs for $L$ steps.
  (b) At depth $t$, $s_t$ softly mixes $E$ expert parameter sets into one dense FFN
  shared by all tokens. (c) Selecting $L$ resamples the depth program on a new grid.
  Teal marks shared parameters and amber marks depth and gating.}
  \label{fig:teaser}
  \vspace{-12bp}
\end{figure*}

In this work, we take the remaining step with \methodname{}, which reuses one Transformer module throughout the encoder (\figref{fig:teaser}). We adapt differentiable weight-space expert merging to full-depth recurrence and \textit{use normalized recurrent depth as its sole conditioning signal}. At depth $t$, a small router maps $s_t=t/(L{-}1)$ to soft coefficients over an $E$-expert FFN bank. Their weighted combination produces one dense FFN that is applied to every token. Across depth, these mixtures trace a continuous path through FFN parameter space, replacing the independently learned FFNs of a standard stack. Changing $L$ resamples the same path on a different grid, separating stored expert capacity from executed depth. Because the schedule is depth (but not image) dependent, a selected deployment depth can also be folded into a fixed sequence of dense FFNs before compilation, eliminating online routing and merging at the cost of materializing one FFN per depth. Under equal compute budget, our depth-programmed merge outperforms the tested token-dispatch and output-mixture alternatives \citep{csordas2024moeut,liu2024deepseek,qwen2025qwen3,wang2025remoe}.

We evaluate \methodname{} with supervised ImageNet-1k training or distillation from a frozen DINOv2 teacher. Trained from scratch, an $E{=}4$ \methodname{}-B/16 matches DeiT~III~\citep{touvron2022deit3} at near-matched inference FLOPs with $73\%$ fewer parameters. Under DINOv2 distillation~\citep{oquab2024dinov2}, an $E{=}8$ model nearly matches the teacher's ImageNet linear-probe accuracy.

Our contributions are:
{\setlength{\leftmargini}{2em}
\setlength{\topsep}{2pt}
\setlength{\partopsep}{0pt}
\setlength{\itemsep}{2pt}
\setlength{\parsep}{0pt}
\setlength{\parskip}{0pt}
\begin{itemize}
  \item \textbf{A recurrent ViT with a continuous depth program:}
    We present \methodname{}, which replaces a ViT's depth-wise stack with one recurrent Transformer module. A normalized depth coordinate programs a continuous trajectory through a shared FFN expert bank, recovering depth-specific computation while remaining competitive with full-depth ViTs.

  \item \textbf{Comparison under a common recurrent setting:}
    Using the same recurrent backbone, task loss, and base training recipe, we compare seven alternative MoE mechanisms. At the nominal compute budget of one dense FFN (per block), our depth-programmed merge outperforms the tested token-dispatch and output-mixture adaptations.

  \item \textbf{Resampleable depth and fixed-depth export:}
  One \methodname{} checkpoint (trained model) supports multiple tested inference depths by resampling the normalized coordinate interval. For deployment, \methodname{} can retain its compact dynamic graph or materialize the depth-specific FFNs at a fixed depth, trading compact storage for conventional dense execution.
 
  \item \textbf{Analysis of the learned depth program:}
Our analysis shows that, under final-layer distillation alone, the router learns where the dominant expert changes and blends experts across each transition, and that the learned gate-expert assignment matters for teacher alignment. Increasing expert capacity improves coverage of the teacher's layers, while larger backbones use fewer effective directions and gain less from additional experts.
  \end{itemize}
}

\section{Closely Related Work}

\noindent\textbf{Recurrent and adaptive depth:} Universal Transformers reuse a transition across depth, optionally with adaptive computation time
\citep{dehghani2019universal,graves2016adaptive}. ALBERT demonstrates the parameter savings of cross-layer sharing in language \citep{lan2020albert}. In vision, Sliced Recursive Transformers and MiniViT share or multiplex weights primarily for compression \citep{shen2022sliced,zhang2022minivit}. Raptor uses $k$ recurrent block templates and, under distillation, intermediate teacher features \citep{jacobs2026block}. Vision-MoR assigns a recursion depth to each patch, while Edge-RecViT combines a shared middle block with token-wise early exit \citep{he2026visionmor,li2026edgerecvit}. Unlike these works, we target a different setting: one block is reused throughout the model, a global depth is selected at inference, and a shared expert bank provides different FFN weights at different depths. When applied, distillation uses only the teacher's final-layer features.

\noindent\textbf{Experts under recurrent sharing:}
Sparse MoEs conventionally dispatch tokens through a subset of expert FFNs \citep{shazeer2017outrageously}. Representative FFN-MoE ViTs place token-routed expert banks inside otherwise untied stacks \citep{riquelme2021scaling}. Sparse Universal Transformer instead combines a fully shared universal layer with sparse token routing \citep{tan2023sparse}. MoEUT recurrently repeats a small group of distinct layers with token-routed experts \citep{csordas2024moeut} while Mixture-of-Recursions routes tokens over recursive depth \citep{bae2025mixture}. Soft MoE forms differentiable token-slot mixtures \citep{puigcerver2024from}. These methods primarily choose computation at token level. In contrast, \methodname{} computes one gate from the normalized coordinate of each recurrent depth. The gate is shared across inputs and tokens, and expert parameters are merged before execution. Our comparisons test how this depth-programmed composition differs from token-dispatch and output-mixture mechanisms under full-depth recurrence.

\noindent\textbf{Conditional parameterization and depth programming:}
Hypernetworks generate one module's parameters from a conditioning signal and include designs based on learned layer codes \citep{ha2017hypernetworks}. Our router is a constrained hypernetwork: it predicts $E$ softmax coefficients over a learned bank of complete FFNs rather than emitting their parameters directly. CondConv, SMEAR, and Lory likewise synthesize mixtures of stored modules using gates driven by image dependent representations \citep{yang2019condconv,muqeeth2023soft,zhong2024lory}. We retain this parameter-composition mechanism but replace the image feature input with normalized recurrent depth as an explicit routing signal, allowing the same router to be evaluated on a recomputed coordinate grid when inference depth changes. FiLM provides a complementary form of conditioning through feature-wise affine modulation of activations \citep{perez2018film}, whereas \methodname{} composes the complete FFN parameters used at each depth.

\section{Method}
\label{sec:method}

The proposed \methodname{} replaces the $L$ independently parameterized blocks of a ViT with one recurrent pre-norm module comprising shared attention and normalization together with a \textit{depth-programmed FFN mixture} (\figref{fig:teaser}).

\subsection{Recurrent block and depth-programmed experts}
\label{sec:method:backbone}
\label{sec:method:merge}

We form $\mX^0\in\R^{T\times d}$ by embedding the image patches, prepending a class token, and adding
learned spatial positional embeddings \citep{dosovitskiy2021image}. For a recurrent depth of $L\geq2$, the
step at depth $t$ is assigned the normalized coordinate $s_t^{(L)}=t/(L-1)$. The router maps this scalar
coordinate to expert logits and mixture weights:
\begin{align}
  \vz_t &= \psi\!\left(s_t^{(L)}\right),
  & \vg_t &= \softmax(\vz_t/\tau),\qquad \tau=1,
  \label{eq:router}
\end{align}
where $t=0,\ldots,L-1$ and $\psi:\R\rightarrow\R^E$ is a two-layer MLP. Because $\vg_t$ depends only
on normalized depth, it is shared by every token and input evaluated at depth $t$. Further implementation
details are in Appendix~\ref{app:implementation}.

The shared bank contains $E$ dense FFNs,
$\theta_e^{\mathrm{ffn}}=(\mW_1^e,\vb_1^e,\mW_2^e,\vb_2^e), \; e=1,\dots,E$. Before the FFN executes, the gate
merges both projections and biases:
\begin{align}
  \bar\theta_t^{\mathrm{ffn}}
    &= \sum_{e=1}^{E}g_{t,e}\theta_e^{\mathrm{ffn}}
     =(\bar{\mW}_{1,t},\bar{\vb}_{1,t},\bar{\mW}_{2,t},\bar{\vb}_{2,t})
     \label{eq:merge-weights}\\
  \operatorname{FFN}_t(\vu)
    &= \bar{\mW}_{2,t}\operatorname{GELU}
       \!\left(\bar{\mW}_{1,t}\vu+\bar{\vb}_{1,t}\right)+\bar{\vb}_{2,t}.
       \label{eq:merge}
\end{align}
At each recurrent depth, the expert parameters are merged into a single FFN, which is then applied to
all tokens. Only the merged FFN is
evaluated, not the individual experts.

With the gate and merged FFN set by $s_t^{(L)}$, one recurrent step is:
\begin{equation}
\begin{aligned}
  \mH^t
    &= \mX^t
      + \operatorname{MHSA}\!\left(\operatorname{LN}_{\mathrm{attn}}(\mX^t)\right),\\
  \mX^{t+1}
    &= \mH^t
      + \operatorname{FFN}_t\!\left(\operatorname{LN}_{\mathrm{ffn}}(\mH^t)\right).
\end{aligned}
\label{eq:recurrence}
\end{equation}
All recurrent steps share the attention, LayerNorms, router, and expert bank. A final LayerNorm and task head then consume $\mX^L$.

Our router takes the normalized coordinate $s_t^{(L)}$ directly rather than assigning a learned embedding to each depth, as in static hypernetworks \citep{ha2017hypernetworks}. Together, Eqs.~\ref{eq:router}-\ref{eq:merge-weights} map any $s\in[0,1]$ to a merged FFN. A model of depth $L$ evaluates this map at $L$ evenly spaced points. Choosing a different $L$ changes only the sampling grid, allowing the same learned depth program to be used without retraining. We evaluate this property in Section~\ref{sec:experiments:anytime}.

Because this map depends only on depth, the merged FFNs are the same for every image. Once $L$ is chosen, they can be computed once and reused or inserted into a conventional fixed-depth graph with no routing or expert merging in the input path. The compact form stores $E$ expert FFNs, whereas the fixed graph stores the $L$ merged FFNs directly. In both cases, each depth evaluates one dense FFN, so increasing $E$ adds stored capacity without increasing dense FFN computation.

\subsection{Training and elastic-depth inference}
\label{sec:method:training}

\noindent\textbf{Task objective:}
Under supervised training, models use $\Ls_{\mathrm{task}}=\Ls_{\mathrm{cls}}$, the same ImageNet classification objective as their untied baselines. For distillation, let $\mY\in\R^{T\times d}$ be the frozen teacher's final post-LayerNorm token tensor. We set $\Ls_{\mathrm{task}}=\Ls_{\mathrm{dist}}$, where
\begin{equation}
  \Ls_{\mathrm{dist}}
  = \frac{1}{Td}\left\lVert\operatorname{LN}(\mX^L)-\mY\right\rVert_F^2,
  \label{eq:distillation-loss}
\end{equation}
is ordinary elementwise mean squared error over all tokens. Unlike \citet{jacobs2026block}, we do not use intermediate feature distillation.

\noindent\textbf{Regularization:}
In order to encourage diversity and improve stability, for every $E>1$ \methodname{} model, we add three auxiliary losses: usage balance, router z-loss, and expert diversity.
Let $\vg_t$ be the gate at recurrent depth $t$.
The balance term is:
\begin{equation}
  \Ls_{\mathrm{bal}}
  = \frac{1}{L}\sum_{t=0}^{L-1}
     \left[
       1-\frac{H(\vg_t)}{\log E}
     \right],
  \label{eq:balance-loss}
\end{equation}
where $H$ denotes Shannon entropy. This term favors soft mixtures of experts.
We also regularize the routing logits and expert parameters: the router z-loss
$\Ls_z$ controls logit scale by penalizing the squared log-sum-exp of the logits,
while the expert-diversity loss $\Ls_{\mathrm{div}}$ encourages distinct expert
weights by penalizing squared cosine similarity between their parameter vectors.
These two losses are defined in Eqs.~\ref{eq:z-loss} and~\ref{eq:diversity-loss}
in Appendix~\ref{app:implementation}. Coefficient sensitivity is reported in
Appendix~\ref{app:training-ablations}. The full objective is:
\begin{equation}
  \Ls = \Ls_{\mathrm{task}}
  +\lambda_{\mathrm{bal}}\Ls_{\mathrm{bal}}
  +\lambda_z\Ls_z+\lambda_{\mathrm{div}}\Ls_{\mathrm{div}}.
  \label{eq:objective}
\end{equation}
For $E=1$, all three expert-specific terms are omitted and the tied-block baseline uses $\Ls_{\mathrm{task}}$ alone.

\noindent\textbf{Elastic-depth training:}
During elastic-depth training, standard per-image DropPath is applied to complete recurrent steps
\citep{huang2016stochastic}. A dropped step acts as the identity
for all tokens in that image. The retained steps are re-numbered, and their coordinates are spread from $0$ to
$1$. This exposes the router to multiple effective depths.

\noindent\textbf{Elastic-depth inference:} At inference, a global depth $L\geq2$ is selected before execution, and the normalized-depth grid is recomputed for its $L$ steps. Changing $L$ changes both the coordinate grid and the recurrence depth. Performance across the tested depths is reported in \secref{sec:experiments:anytime}.

\section{Experiments}
\label{sec:experiments}

We study three questions: whether one recurrent block can replace a deep ViT under supervised training and DINOv2 distillation, which expert-composition mechanisms remain effective under full-depth reuse, and whether a single checkpoint (trained model) can support multiple inference depths. We also vary the stored expert-bank size.

\subsection{Experimental setup}
\label{sec:experiments:setup}

We consider two regimes: ImageNet-1k training from scratch and distillation from a frozen DINOv2 teacher, evaluated with frozen-backbone probes. All models were implemented in PyTorch. For additional implementation details see Appendix~\ref{app:implementation}. 

Unless stated otherwise, the tables report checkpoints evaluated at their reference depth.
\Figref{fig:anytime-depth} evaluates the distilled routing variants across inference depths.
For the supervised comparison, \methodname{}, DeiT~III, and Raptor use the same data, base training recipe, recurrent depth, and near-matched inference FLOPs. The distilled comparison uses different objectives: \methodname{} matches only final-layer teacher features, while Raptor was trained with intermediate features.

\noindent\textbf{Supervised ImageNet-1k training:}
We train \methodname-S, -B, and -L at their native ViT depths using a shortened DeiT~III recipe \citep{touvron2022deit3}, sweeping $E\in\{1,2,3,4,8\}$. For $E{=}1$, the model reduces to a single Transformer block reused at every recurrent step. 
\noindent\textbf{Distillation from DINOv2:}
For comparison with \citet{jacobs2026block}, we distill a \methodname{}-B/14 student from a frozen DINOv2-B/14 teacher \citep{oquab2024dinov2}. The key difference is the supervision target: Raptor matches intermediate teacher features, whereas our loss compares only the final student and teacher features. We evaluate the frozen student using an ImageNet-1k linear classifier, an ADE20k linear segmentation head, and linear and two-layer MLP depth heads on NYUv2.

\subsection{Can one recurrent block replace a deep ViT?}
\label{sec:experiments:main}

\noindent\textbf{Supervised training:}
At near-matched inference FLOPs, Table~\ref{tab:main-scratch} shows that the $E{=}4$ \methodname{} models score $0.1$--$0.2$ points above four-block Raptor across all scales. Relative to full-depth DeiT~III, \methodname-B/16 reaches $83.0\%$ versus $82.8\%$, and \methodname-L/16 reaches $83.8\%$ versus $84.1\%$ with roughly $7\times$ fewer parameters. The S/16 gap is larger ($78.2\%$ versus $79.9\%$), but narrows with additional experts
(Table~\ref{tab:experts}). These results show that \methodname's shared module matches or closely approaches full-depth ViTs at larger scales and outperforms four-block Raptor at every scale.

\begin{table}[t]
\centering
\caption{\textbf{ImageNet-1k training from scratch.} All rows use the same recipe, and \methodname{} values are means over three runs. Recurrent depths match the corresponding ViTs. $k$ - counts distinct block templates and $E$ - num. experts.}
\vspace{-6bp}
\label{tab:main-scratch}
\small
\setlength{\tabcolsep}{4.5pt}
\begin{tabular}{llccrrr}
\toprule
Scale & Method & $k$ & $E$ & Params (M) & GFLOPs/image & Top-1 (\%) $\uparrow$ \\
\midrule
\multirow{3}{*}{S/16}
 & DeiT~III \citep{touvron2022deit3} & 12 & -- & 22.1 & 8.5 & 79.9 \\
 & Raptor \citep{jacobs2026block}     & 4 & -- & 7.9 & 8.5 & 78.0 \\
 & \methodname{} (ours)               & 1 & 4 & 6.1 & 8.6 & 78.2 \\
\midrule
\multirow{3}{*}{B/16}
 & DeiT~III                            & 12 & -- & 86.6 & 33.9 & 82.8 \\
 & Raptor                              & 4 & -- & 29.9 & 33.9 & 82.8 \\
 & \methodname{} (ours)               & 1 & 4 & 23.6 & 34.3 & 83.0 \\
\midrule
\multirow{3}{*}{L/16}
 & DeiT~III                            & 24 & -- & 304.4 & 119.9 & 84.1 \\
 & Raptor                              & 4 & -- & 52.4 & 119.9 & 83.7 \\
 & \methodname{} (ours)               & 1 & 4 & 40.9 & 121.5 & 83.8 \\
\bottomrule
\end{tabular}
\vspace{-12bp}
\end{table}

\noindent\textbf{Distillation and frozen transfer:}
Table~\ref{tab:raptor-dinov2} places our distilled models alongside Raptor and the DINOv2 teacher. At $E{=}k\in\{2,3,4\}$, \methodname{} has better reported values on all four probes. These are reference comparisons rather than controlled ablations because the distillation objectives differ. The unmatched (no Raptor model for this case)  $E{=}8$ model is compared only with its teacher. 
\begin{table}[ht]
\centering
\caption{\textbf{DINOv2 distillation and frozen-probe transfer.} \methodname{} values are means over three runs. \methodname{} uses final-layer supervision and Raptor intermediate features. Raptor IN-1k and ADE20k values are published, while NYUv2 values are our checkpoint reevaluations. $E{=}k$ aligns stored FFN count, but not training. The $E{=}8$ row is compared only with DINOv2-B/14.}
\label{tab:raptor-dinov2}
\vspace{-6bp}
\small
\setlength{\tabcolsep}{6pt}
\begin{tabular}{lcrrrr}
\toprule
Method & Config & IN-1k & ADE20k & NYUv2 (lin) & NYUv2 (MLP) \\
       &        & {\footnotesize Top-1\,\%\,$\uparrow$} & {\footnotesize mIoU\,\%\,$\uparrow$} & {\footnotesize RMSE\,(m)\,$\downarrow$} & {\footnotesize RMSE\,(m)\,$\downarrow$} \\
\midrule
\methodname~(ours) & $E{=}1$ & 79.4 & 39.7 & 0.583 & 0.529 \\
\midrule
\methodname~(ours) & $E{=}2$ & 82.2 & 43.0 & 0.559 & 0.498 \\
Raptor~\citep{jacobs2026block} & $k{=}2$ & 81.2 & 39.6 & 0.605 & 0.548 \\
\midrule
\methodname~(ours) & $E{=}3$ & 83.1 & 43.6 & 0.555 & 0.491 \\
Raptor & $k{=}3$ & 83.0 & 43.0 & 0.567 & 0.506 \\
\midrule
\methodname~(ours) & $E{=}4$ & 83.5 & 44.6 & 0.552 & 0.491 \\
Raptor & $k{=}4$ & 83.2 & 43.6 & 0.564 & 0.495 \\
\midrule
\methodname~(ours) & $E{=}8$ & 83.9 & 45.8 & 0.554 & 0.483 \\
\midrule
\multirow{2}{*}{\gr{\textit{DINOv2}~\citep{oquab2024dinov2}}}
  & \gr{\textit{S/14}} & \gr{80.9} & \gr{44.6} & \gr{0.590} & \gr{0.527} \\
  & \gr{\textit{B/14}} & \gr{84.5} & \gr{47.5} & \gr{0.567} & \gr{0.486} \\
\bottomrule
\end{tabular}
\vspace{-6bp}
\end{table}

\subsection{Which MoE formulation suits a single recurrent block?}
\label{sec:moe-baselines}

Using multiple FFN experts improves accuracy over a single shared FFN in both training regimes (Tables~\ref{tab:raptor-dinov2} and~\ref{tab:experts}). We next ask a narrower question: which expert-composition mechanisms remain effective when adapted to a single Transformer block reused at every depth? We instantiate each mechanism in the same recurrent ViT, holding the data, backbone, recurrent depth, task objective, and base recipe fixed. The comparison measures compatibility with full-depth recurrence in a common testbed, rather than ranking the original systems in their native architectures.

We compare the FFN computation used by each method per recurrent step,
taking one standard dense FFN as the reference. The ratio $U$ measures
nominal FFN FLOPs relative to this reference: $U{=}1$ matches its computation,
while $U{=}4$ uses four times as much. Methods with the same $U$ may still differ in storage and end-to-end execution cost. At $U{=}1$, we ask which adaptation best recovers the accuracy lost through full-depth sharing without adding FFN computation. Weight-space merging evaluates one dense FFN regardless of $E$, so adding experts increases stored capacity while $U$ remains unchanged. To obtain $U{>}1$ would require a wider merged FFN or multiple FFN evaluations, which scales dense computation rather than the expert bank itself. We therefore report weight-merge methods at $U{=}1$ and study $E$ separately in Section~\ref{sec:experiments:experts}.

Because the original methods are designed around different expert widths and activation patterns, increasing $U$ requires method-specific changes. The higher-$U$ configurations therefore test whether relaxing the one-FFN constraint closes the gap for each method, rather than tracing a common scaling curve. Appendix~\ref{app:moe-baselines} reports expert-FFN storage and documents each adaptation.

\noindent\textbf{Supervised training.}
At the primary $U{=}1$ budget, \methodname{} reaches $78.2\%$, Lory $77.9\%$, and SMEAR $77.7\%$, while every other method falls between $67.8\%$ and $70.6\%$ (Table~\ref{tab:moe-baselines}). Several alternatives narrow this gap at higher nominal budgets, with MoEUT reaching $77.6\%$ at $U{=}4$.

\begin{table}[t]
\centering
\caption{\textbf{MoE formulations under single-block recurrence: supervised ImageNet-1k training.} All methods use the same -S backbone, $L{=}12$, and base recipe, only the MoE mechanism changes. Higher-$U$ settings also change expert count, activation count, and/or width.}
\label{tab:moe-baselines}
\vspace{-6bp}
\small
\setlength{\tabcolsep}{5pt}
\renewcommand{\arraystretch}{1.05}
\begin{tabular}{@{}l l cccc@{}}
\toprule
& & \multicolumn{4}{c}{Top-1 (\%) $\uparrow$ at nominal budget $U$} \\
\cmidrule(lr){3-6}
Family & Method & $1\times$ & $1.5\times$ & $2\times$ & $4\times$ \\
\midrule
No MoE & \methodname{} ($E{=}1$) & 70.6 & -- & -- & -- \\
\midrule
\multirow{3}{*}{Weight merge} & \methodname{} (ours) & \textbf{78.2} & -- & -- & -- \\
 & Lory~\citep{zhong2024lory} & 77.9 & -- & -- & -- \\
 & SMEAR~\citep{muqeeth2023soft} & 77.7 & -- & -- & -- \\
\midrule
\multirow{3}{*}{Token dispatch} & DeepSeek-V3~\citep{liu2024deepseek} & 70.6 & 71.3 & 72.4 & 76.5 \\
 & Qwen3~\citep{qwen2025qwen3} & 68.6 & 71.0 & 72.5 & 74.0 \\
 & MoEUT~\citep{csordas2024moeut} & 67.8 & 71.4 & 72.8 & \textbf{77.6} \\
\midrule
\multirow{2}{*}{Output mixture} & ReMoE~\citep{wang2025remoe} & 70.1 & 71.9 & 72.8 & 75.5 \\
 & Soft-MoE~\citep{puigcerver2024from} & 69.2 & \textbf{72.3} & \textbf{73.7} & 75.7 \\
\bottomrule
\end{tabular}
\vspace{-0.3cm}
\end{table}

\noindent\textbf{Distillation.}
Distillation preserves the main $U{=}1$ result, although the relative strength of the baselines changes. At $U{=}1$, \methodname{} has the best reported ImageNet, ADE20k, and linear-depth results, while SMEAR achieves the lower MLP-head RMSE ($0.485$ versus $0.491$) and otherwise remains close to \methodname{} (Table~\ref{tab:moe-dino}). The \methodname{} row is the $E{=}4$ model from Table~\ref{tab:raptor-dinov2}.  The Qwen3 run at $U{=}1$ did not converge to a usable solution. At $U{=}4$, ReMoE leads on all four tasks, although these higher-budget configurations also change the number or width of the active experts. Additional configurations are reported in Appendix~\ref{app:moe-baselines}. Across both training regimes, the depth-programmed merge provides the strongest overall results at the primary $U{=}1$ budget, while other formulations become competitive when given greater active FFN capacity.

\begin{table}[t]
\centering
\caption{\textbf{MoE formulations under single-block recurrence: DINOv2 distillation.} All entries use the same -B backbone, $L{=}12$, and task loss. Each cell gives $U{=}1/U{=}4$. Additional results are shown in Appendix~\ref{app:moe-baselines} (Table~\ref{tab:moe-dino-full}).}
\label{tab:moe-dino}
\vspace{-6bp}
\footnotesize
\setlength{\tabcolsep}{2.75pt}
\renewcommand{\arraystretch}{1.05}
\begin{tabular}{@{}llcccc@{}}
\toprule
Family & Method & IN-1k $\uparrow$ & ADE20k $\uparrow$ & NYUv2 lin. $\downarrow$ & NYUv2 MLP $\downarrow$ \\
\midrule
No MoE & \methodname{} ($E{=}1$) & 79.4 / -- & 39.7 / -- & 0.583 / -- & 0.529 / -- \\
\midrule
\multirow{3}{*}{Weight merge}
 & \methodname{} ($E{=}4$) & \textbf{83.5} / -- & \textbf{44.6} / -- & \textbf{0.552} / -- & 0.491 / -- \\
 & Lory~\citep{zhong2024lory} & 79.2 / -- & 39.9 / -- & 0.580 / -- & 0.531 / -- \\
 & SMEAR~\citep{muqeeth2023soft} & 83.4 / -- & 44.3 / -- & 0.553 / -- & \textbf{0.485} / -- \\
\midrule
\multirow{3}{*}{Token dispatch}
 & DeepSeek-V3~\citep{liu2024deepseek} & 74.4 / 76.6 & 36.6 / 39.9 & 0.654 / 0.607 & 0.582 / 0.539 \\
 & Qwen3~\citep{qwen2025qwen3} & -- / 74.9 & -- / 40.1 & -- / 0.638 & -- / 0.550 \\
 & MoEUT~\citep{csordas2024moeut} & 75.1 / 83.0 & 35.0 / 43.5 & 0.694 / 0.558 & 0.608 / 0.501 \\
\midrule
\multirow{2}{*}{Output mixture}
 & ReMoE~\citep{wang2025remoe} & 79.6 / \textbf{83.2} & 39.7 / \textbf{45.1} & 0.573 / \textbf{0.539} & 0.527 / \textbf{0.482} \\
 & Soft-MoE~\citep{puigcerver2024from} & 78.5 / 82.2 & 33.4 / 37.1 & 0.602 / 0.547 & 0.557 / 0.500 \\
\bottomrule
\end{tabular}
\vspace{-12bp}
\end{table}

\subsection{Depth-programmed experts enable elastic-depth inference}
\label{sec:experiments:anytime}

\begin{figure}[ht]
\centering
\includegraphics[width=0.9\linewidth]{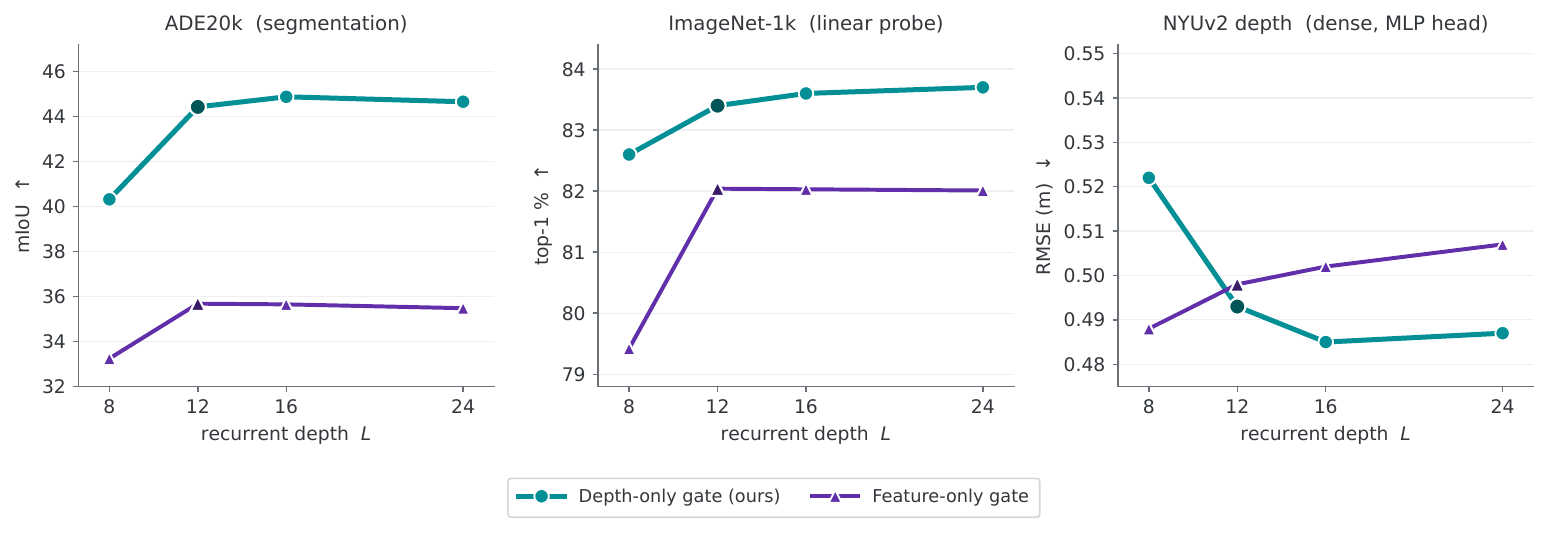}
\vspace{-10bp}
\caption{\textbf{Routing variants under elastic-depth training.} Both variants use the same
per-image depth-dropout protocol and are evaluated at $L\in\{8,12,16,24\}$.}
\label{fig:anytime-depth}
\vspace{-0.3cm}
\end{figure}

\Figref{fig:anytime-depth} asks whether a single model can support multiple inference depths.
We compare two distilled variants \textit{trained with the same per-image depth dropout}: our
depth-only gate and a SMEAR-style feature-only gate with no explicit depth signal.
At $L{=}12$, the feature-only control reaches approximately $35.6$ ADE20k mIoU. Our depth-only variant
reaches $44.4$ mIoU, $83.4\%$ ImageNet linear-probe top-1, and $0.493$ NYUv2 RMSE.

Our proposed variant improves on all three probes through $L{=}16$ and remains near its best at $L{=}24$. ADE20k rises from $40.3$ mIoU at $L{=}8$ to $44.9$ at $L{=}16$ and remains at $44.7$ at $L{=}24$. NYUv2 RMSE improves from $0.522$ to $0.485$ and remains close at $0.487$, while ImageNet top-1 increases from $82.6\%$ to $83.7\%$. Beyond $L{=}12$, the feature-only control is nearly flat on ImageNet and ADE20k. It also worsens on NYUv2, from $0.498$ at $L{=}12$ to $0.507$ at $L{=}24$. Normalized-depth conditioning therefore preserves reference-depth performance during training and converts additional recurrent steps into gains through $L{=}16$.

Together with \secref{sec:moe-baselines}, these results show that weight-space merging is effective at the
primary compute budget, while normalized-depth conditioning preserves accuracy across the tested depths
and benefits from greater depth.

\subsection{Number of experts}
\label{sec:experiments:experts}
\label{sec:experiments:ablations}

Expanding the merged expert bank consistently recovers accuracy lost under plain weight tying while retaining one dense FFN evaluation per step. From $E{=}1$ to $E{=}8$, ImageNet-1k top-1 increases by $9.6$, $6.8$, and $5.7$ points for S/16, B/16, and L/16, respectively (Table~\ref{tab:experts}). The largest single improvement at every scale occurs from $E{=}1$ to $E{=}2$. Beyond $E{=}4$, additional capacity primarily benefits S/16. 

\begin{table}[ht]
 \caption{\textbf{Effect of expert-bank size on ImageNet-1k.} All models use supervised training.}
\label{tab:experts}
\vspace{-8bp}
\centering
\begin{tabular}{lccccccc}
\toprule
& & \multicolumn{5}{c}{Number of experts $E$} & \\
\cmidrule(lr){3-7}
Model & DeiT~III & $E{=}1$ & $E{=}2$ & $E{=}3$ & $E{=}4$ & $E{=}8$ & $\Delta_{1\rightarrow8}$ \\
\midrule
\methodname-S/16 & 79.9 & 70.6 & 74.2 & 76.5 & 78.2 & \textbf{80.2} & $+9.6$ \\
\methodname-B/16 & 82.8 & 76.4 & 81.9 & 82.5 & 83.0 & \textbf{83.2} & $+6.8$ \\
\methodname-L/16 & 84.1 & 78.3 & 83.0 & 83.5 & 83.8 & \textbf{84.0} & $+5.7$ \\
\bottomrule
\end{tabular}
\vspace{-0.3cm}
\end{table}

\section{Analysis: what does the recurrent block learn?}
\label{sec:analysis}

We examine how training organizes the expert bank across depth and whether the learned assignment matters
for the final representation. We then relate expert utilization to gains from larger banks across model
scales. Section~\ref{sec:analysis:routing} analyzes distilled \methodname{}-B/14, while
Section~\ref{sec:analysis:capacity} compares supervised S/16 and L/16 models. Appendix~\ref{app:analysis}
provides the protocols, results, and expert-ablation controls.

\subsection{How the router allocates experts over depth}
\label{sec:analysis:routing}

\begin{figure}[!ht]
  \centering
  \begin{minipage}[t]{0.45\linewidth}
    \centering
    \includegraphics[width=\linewidth]{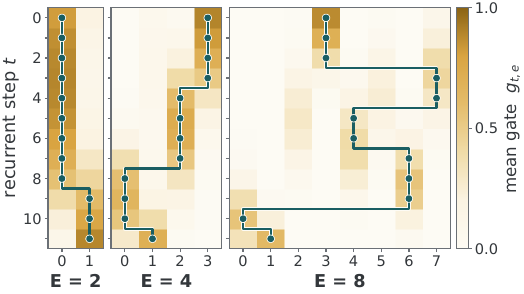}
    \vspace{-18bp}
    \captionsetup{hypcap=false}
    \captionof{figure}{\textbf{Expert routing across recurrent depth.} Gate weights $g_{t,e}$ for distilled \methodname-B/14 ($L{=}12$). The teal path marks the expert with the largest weight at each depth, showing how the dominant expert changes over the recurrence.}
    \label{fig:gates}
  \end{minipage}
  \hfill
  \begin{minipage}[t]{0.53\linewidth}
    \centering
    \includegraphics[width=\linewidth]{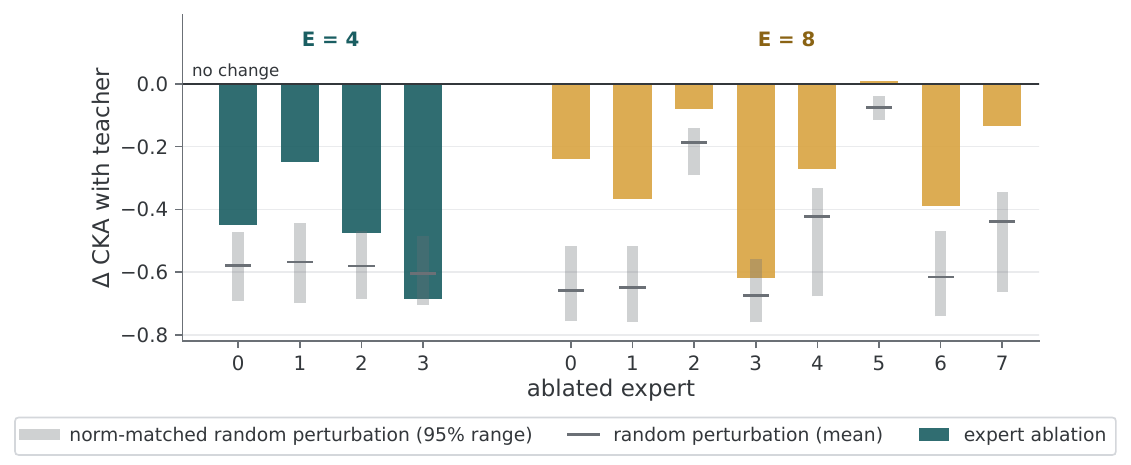}
    \vspace{-18bp}
    \captionsetup{hypcap=false}
    \captionof{figure}{\textbf{Expert ablation against matched controls.} Change in teacher CKA after ablating each expert. Gray ranges and ticks show the middle $95\%$ and mean of 64 random gate perturbations matched in merged-weight displacement.}
    \label{fig:drop}
  \end{minipage}
  \vspace{-0.2cm}
\end{figure}

We first examine how each expert contributes across depth. In \figref{fig:gates},
the dominant expert changes only a few times, with each dominant expert occupying a contiguous depth interval.
The gate remains a soft mixture, with overlapping expert weights around the transitions.
The mixtures are visibly more diffuse at $E{=}8$. No objective specifies these intervals or their boundaries.
Their organization resembles Raptor's predefined recurrent segments \citep{jacobs2026block}, but here the
only teacher targets are the final-layer features. The continuous router also defines mixtures between the
plotted depths, allowing the learned schedule to be sampled at a different inference depth.

We next measure how removing each expert affects the final representation. We set its gate weight to zero
at every depth, renormalize the remaining weights, and measure the linear centered kernel alignment \citep[CKA,][]{kornblith2019similarity} between the student's and teacher's final features. The resulting CKA drop
reflects both how much the merged weights change and where those changes occur in the recurrence, making it
difficult to isolate the contribution of the deleted expert. Figure~\ref{fig:drop} compares each deletion
with 64 random gate perturbations matched in their displacement of the merged FFN
weights. Most deletions reduce teacher alignment, but less than random perturbations of the same size.
Removing expert 5 at $E{=}8$ does not reduce CKA. The largest losses occur when deleting the earliest-used
experts (Appendix~\ref{app:analysis}), so deletion sensitivity alone cannot separate an expert's role from
its position in the recurrence.

To test whether experts can exchange assignments, we permute the gate columns while keeping
the expert bank fixed. This preserves every expert and the mixing coefficients at each depth, changing only
which expert receives each coefficient. All 23 nonidentity permutations at $E{=}4$ and all 64 tested
permutations at $E{=}8$ reduce teacher CKA (Appendix~\ref{app:analysis}). Thus, retaining the full bank
is not sufficient to preserve teacher alignment: the learned assignment of experts across depth also matters.

\subsection{How fully is the expert bank used?}
\label{sec:analysis:capacity}

The benefit of a larger expert bank varies with model scale. Increasing $E$ from four to eight improves
S/16 by $2.0$ ImageNet points but L/16 by only $0.2$ (Table~\ref{tab:experts}). We therefore ask whether
the two models differ in how fully they utilize the available expert directions.

\begin{figure}[!ht]
  \centering
  \begin{minipage}[t]{0.435\linewidth}
    \centering
    \includegraphics[width=\linewidth]{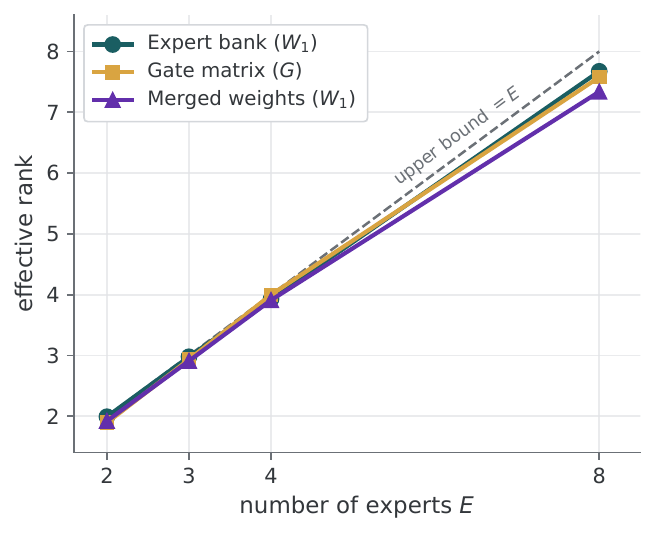}\\[1pt]
    {\bf (a)} \methodname-S/16 ($L{=}12$)
  \end{minipage}
  \hfill
  \begin{minipage}[t]{0.435\linewidth}
    \centering
    \includegraphics[width=\linewidth]{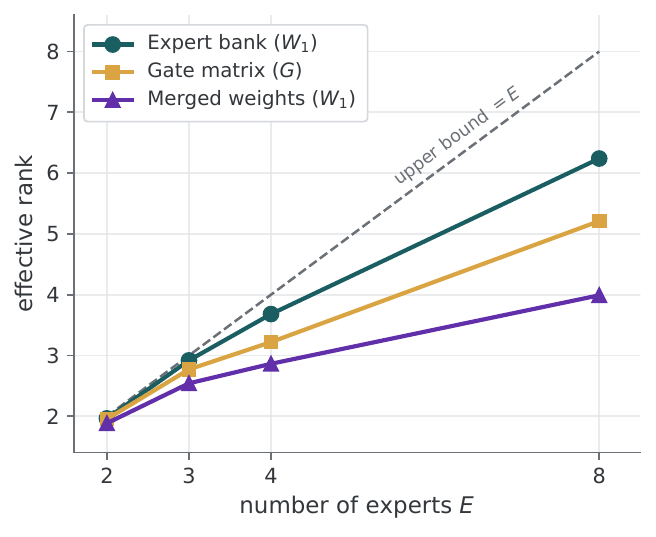}\\[1pt]
    {\bf (b)} \methodname-L/16 ($L{=}24$)
  \end{minipage}
  \vspace{-6bp}
  \caption{\textbf{Available, selected, and realized expert directions.}
  Effective ranks of the stacked expert $W_1$ matrices, gate weights across depth, and stacked merged
  $W_1$ matrices. }
  \label{fig:rank}
  \vspace{-12bp}
\end{figure}

A diverse expert bank can still produce similar merged weights across depth if the router repeatedly
selects similar mixtures. Fig.~\ref{fig:rank} therefore compares the expert bank, routing schedule,
and merged weights. We stack the flattened expert projections $\mW_1^e$ into
$\mW_{1,\mathrm{bank}}\in\R^{E\times P}$, where $P=d_{\mathrm{ff}}d$, with model width $d$
and FFN hidden width $d_{\mathrm{ff}}$.
The routing schedule forms $\mG\in\R^{L\times E}$, with entries $G_{t,e}=g_{t,e}$ and one
row per recurrent depth. The corresponding merged weights form
$\mW_{1,\mathrm{traj}}=\mG\mW_{1,\mathrm{bank}}\in\R^{L\times P}$, whose rows are the
flattened merged projections $\operatorname{vec}(\bar{\mW}_{1,t})^\top$.
Their effective ranks measure, respectively, the diversity available in the expert bank,
the mixtures selected across depth, and the weights actually used by the recurrent block.

We compute effective rank as the exponential of the entropy of the normalized squared singular values
(Appendix~\ref{app:analysis}). It approaches one when a single direction dominates and equals $k$ when
$k$ nonzero singular values are equal. Because the diversity loss encourages the stored experts to differ,
the gate and merged-weight ranks show whether that diversity is also expressed during execution.
At $E{=}8$, the expert, gate, and merged-weight ranks are $7.7$, $7.6$, and $7.3$ for S/16, indicating
that almost all available directions are realized. For L/16, the corresponding ranks are $6.2$, $5.2$, and
$4.0$, indicating that its merged weights are concentrated in fewer directions. This matches the accuracy
trend: S/16 gains more from a larger expert bank and makes broader use of its available weight directions
across depth. The same pattern remains after removing
the mean weight and gate vectors across depth (Appendix~\ref{app:analysis}).

\FloatBarrier

\FloatBarrier
\section{Conclusion}

\methodname{} replaces a ViT's depth-wise stack with one recurrent Transformer module. A continuous
normalized-depth coordinate softly merges a shared expert bank into one FFN at each recurrent depth. This
learned parameter trajectory recovers much of the accuracy lost under plain weight tying while retaining
substantially fewer stored parameters than a full-depth ViT. At matched one-FFN compute, it outperforms the
tested token-dispatch and output-mixture alternatives. Elastic-depth training allows the same trajectory
to be sampled at multiple tested depths. For deployment, \methodname{} can retain its compact dynamic graph
or materialize the depth-specific FFNs at a fixed depth, trading compact storage for conventional dense execution.

\beginappendix
\raggedbottom
\makeatletter
\setlength{\@fptop}{0pt}
\makeatother

\section{Implementation, training, and evaluation details}
\label{app:implementation}

\subsection{Model implementation}

\paragraph{Router and expert bank:}
The routing computation is defined in Eq.~\ref{eq:router}. The router $\psi$ is a two-layer MLP with 16 hidden units and SiLU activation that maps
one normalized-depth coordinate to $E$ logits. Its output layer is zero-initialized, so all experts start with
equal weight. It has no learned step embedding and fixes $\tau{=}1$. The gate therefore depends only on normalized recurrent depth.

Each expert stores
$\mW_1^e\in\R^{d_{\mathrm{ff}}\times d}$,
$\vb_1^e\in\R^{d_{\mathrm{ff}}}$,
$\mW_2^e\in\R^{d\times d_{\mathrm{ff}}}$, and
$\vb_2^e\in\R^d$. Experts are initialized independently using the same initialization as a dense ViT
FFN, with no additional variance correction for the merged weights.

\subsection{Objectives and regularization}

\paragraph{Auxiliary losses:}
In addition to the balance loss in Eq.~\ref{eq:balance-loss}, we use the router z-loss
\citep{zoph2022stmoe}. For router logits $\vz_t$ at recurrent depth $t$:
\begin{equation}
  \Ls_z
  = \frac{1}{L}\sum_{t=0}^{L-1}
    \left(\log\sum_{e=1}^{E}\exp z_{t,e}\right)^2,
  \label{eq:z-loss}
\end{equation}
which controls logit scale during mixed-precision training. We also penalize similarity between expert
parameters. Let $\vq_e=\operatorname{vec}(\theta_e^{\mathrm{ffn}})$ and define
\begin{equation}
  \Ls_{\mathrm{div}}
  = \frac{2}{E(E-1)}\sum_{1\leq e<e'\leq E}
    \left(
      \frac{\vq_e^\top\vq_{e'}}
      {\lVert\vq_e\rVert_2\lVert\vq_{e'}\rVert_2}
    \right)^2.
  \label{eq:diversity-loss}
\end{equation}
We use $\lambda_{\mathrm{bal}}{=}10^{-2}$,
$\lambda_z{=}10^{-3}$, and $\lambda_{\mathrm{div}}{=}10^{-3}$ for $E>1$. All three expert-specific
terms are omitted for $E=1$.

\paragraph{Elastic-depth training.}
For \figref{fig:anytime-depth}, we use per-image depth dropout with rate $0.3$.
DropPath masks complete recurrent steps independently for each image, with one mask shared by all tokens.
A dropped step acts as the identity. Retained steps are reindexed in execution order, and their
normalized-depth coordinates are recomputed from the number retained.

\subsection{Training and evaluation protocols}

\paragraph{Supervised ImageNet-1k training:}
We adopt the ImageNet-1k DeiT~III recipe \citep{touvron2022deit3} with a $300$-epoch schedule.
\methodname{}, DeiT~III, and Raptor \citep{jacobs2026block} use the same data, augmentation, optimizer,
and epoch schedule. Each recurrent model is unrolled to the depth of the standard ViT at its scale:
$L{=}12$ steps for S/16 and B/16, and $L{=}24$ for L/16. Table~\ref{tab:hyperparameters} gives the shared
and scale-specific settings.

\begin{table}[ht!]
  \caption{\textbf{Supervised ImageNet-1k configuration.} The 300-epoch DeiT~III base recipe is shared by
all methods in Table~\ref{tab:main-scratch}. Columns give scale-specific \methodname{} settings, including
recurrence and auxiliary losses.}
\label{tab:hyperparameters}
\centering
\small
\setlength{\tabcolsep}{5pt}
\begin{tabular}{lccc}
\toprule
 & \methodname-S/16 & \methodname-B/16 & \methodname-L/16 \\
\midrule
\multicolumn{4}{@{}l}{\textit{Optimization and schedule}} \\
Optimizer & \multicolumn{3}{c}{AdamW} \\
Peak learning rate & $4\times10^{-3}$ & $3\times10^{-3}$ & $3\times10^{-3}$ \\
Effective batch size & \multicolumn{3}{c}{$2048$} \\
Schedule & \multicolumn{3}{c}{cosine} \\
Epochs & \multicolumn{3}{c}{$300$} \\
Warmup epochs & \multicolumn{3}{c}{$5$} \\
Warmup learning rate & \multicolumn{3}{c}{$10^{-6}$} \\
Weight decay & \multicolumn{3}{c}{$0.05$} \\
\multicolumn{4}{@{}l}{\textit{Loss and augmentation}} \\
Loss & \multicolumn{3}{c}{binary cross-entropy} \\
Label smoothing & \multicolumn{3}{c}{$0.0$} \\
3-Augment & \multicolumn{3}{c}{yes} \\
Color jitter & \multicolumn{3}{c}{$0.3$} \\
Mixup / CutMix & \multicolumn{3}{c}{$0.8$ / $1.0$} \\
Repeated augmentation & \multicolumn{3}{c}{yes} \\
Random erasing & \multicolumn{3}{c}{$0.0$} \\
\midrule
\multicolumn{4}{@{}l}{\textit{Resolution}} \\
Training resolution & \multicolumn{3}{c}{$224$} \\
Evaluation resolution & \multicolumn{3}{c}{$224$} \\
Evaluation crop ratio & \multicolumn{3}{c}{$1.0$} \\
\midrule
\multicolumn{4}{@{}l}{\textit{Recurrence and expert bank}} \\
Recurrent steps $L$ & $12$ & $12$ & $24$ \\
Experts $E$ & \multicolumn{3}{c}{$4$ (sweep: $\{1,2,3,4,8\}$)} \\
Usage balance $\lambda_{\mathrm{bal}}$ & \multicolumn{3}{c}{$10^{-2}$} \\
Router z-loss $\lambda_z$ & \multicolumn{3}{c}{$10^{-3}$} \\
Expert diversity $\lambda_{\mathrm{div}}$ & \multicolumn{3}{c}{$10^{-3}$} \\
\bottomrule
\end{tabular}
\end{table}

\paragraph{DINOv2 distillation:}
The fixed-depth distilled checkpoints use a single-stage schedule at $L{=}12$. Throughout training, a \methodname{}-B/14 student is trained with Eq.~\ref{eq:distillation-loss} to match the final post-LayerNorm representation of a frozen
DINOv2-B/14 teacher \citep{oquab2024dinov2}. Unless stated otherwise, fixed-depth distillation uses the
complete supervised B-scale optimization (see Table~\ref{tab:hyperparameters}). Only the patch size and
task objective differ.
The variants in \figref{fig:anytime-depth} subsequently undergo the elastic-depth training described above.

\paragraph{Frozen-backbone evaluation:}
The distilled encoder remains frozen for all downstream evaluations. We train an ImageNet-1k linear
classifier, an ADE20k linear segmentation head, and linear and two-layer MLP depth heads on NYUv2.
All frozen-backbone probes follow the optimization and evaluation protocol of \citet{jacobs2026block},
applied identically to \methodname{} and the re-evaluated Raptor checkpoints.
The linear depth head measures information available to a linear readout, while the MLP measures what
a shallow nonlinear decoder can recover.

\paragraph{Result provenance:}
All \methodname{} results in Tables~\ref{tab:main-scratch} and~\ref{tab:raptor-dinov2} are means over
three independently trained runs. Downstream probes use fixed random seeds. All rows of
Table~\ref{tab:main-scratch} are our runs under the shared $300$-epoch recipe. The
reproduced DeiT~III results match its published $300$-epoch results. In Table~\ref{tab:raptor-dinov2},
the Raptor ImageNet-1k and ADE20k values are published three-seed means~\citep{jacobs2026block}.
The NYUv2 values are our re-evaluations of its checkpoints using the same probes as \methodname{}.
Raptor uses intermediate teacher features, whereas \methodname{} uses only final-layer features.

\subsection{Computational accounting}

\paragraph{Training resources:}
We train each model on eight NVIDIA H100 GPUs. Depending on model scale, a complete training run takes
roughly 1--2 days, or approximately 192--384 H100 GPU-hours.

\subsection{Measured inference latency}
\label{app:latency}

Table~\ref{tab:h100-latency} reports compiled PyTorch inference for ViT-S models at
$224{\times}224$ resolution in bfloat16 on one NVIDIA H100. Dynamic \methodname{} uses a
$3.6\times$ smaller deployment graph and reduces total runtime memory by $48\%$ at batch 1, while
incurring a $13\%$ latency overhead. At batch 64, the latency overhead rises to $39\%$ and total memory
is similar because batch-dependent allocations dominate. These measurements use the direct PyTorch
implementation, which reconstructs the depth-specific weights using generic FP32 reductions without
cross-input caching or a custom kernel. They therefore characterize the current implementation rather
than an optimized deployment. When $L$ is fixed, the input-independent merged weights can instead be
precomputed, trading the compact runtime representation for conventional dense execution.

\begin{table}[ht]
  \caption{\textbf{Compiled H100 inference.} ViT-S models at $224{\times}224$ resolution in bfloat16.
  Deployment parameters, latency, throughput, and total runtime memory are shown for batches 1 and 64.}
  \label{tab:h100-latency}
  \centering
  \small
  \setlength{\tabcolsep}{2.5pt}
  \begin{tabular}{lcccc}
    \toprule
    Model & Params (M) & Latency (ms) & Throughput (img/s) & Memory (MiB) \\
          &            & B1 / B64     & B1 / B64           & B1 / B64 \\
    \midrule
    DeiT~III-S/16                 & 22.1 & 0.69 / 5.22 & 1,449 / 12,268 & 120.4 / 301.0 \\
    \methodname{}-S/16 (dynamic) & 6.1  & 0.78 / 7.26 & 1,282 / 8,814  & 63.1 / 292.4 \\
    \bottomrule
  \end{tabular}
\end{table}

\FloatBarrier
\section{MoE baseline adaptations}
\label{app:moe-baselines}

Table~\ref{tab:moe-baselines} compares methods using $U$, the nominal expert-FFN computation per step
relative to a standard ViT FFN, whose hidden layer has four times the model width. Thus, $U{=}1$ matches the arithmetic of one dense FFN.
Our primary comparison is at $U{=}1$. Matching $U$ does not match parameter count, total FLOPs, or latency because the higher-budget variants
can change the number, width, or activation pattern of their experts.

We write $E$ for the number of stored experts, $K$ for the number activated per token, and
$K_{\mathrm{sh}}$ for the number of always-active shared experts. The expert hidden ratio
$\mathrm{ehr}$ is the expert width divided by the model width, with $\mathrm{ehr}{=}4$ for a standard
ViT FFN. Token-dispatch methods use $U{=}(K{+}K_{\mathrm{sh}})\mathrm{ehr}/4$. Output mixtures use
$U{=}E\mathrm{ehr}/4$, and Soft-MoE uses
$U{=}(N_{\mathrm{slots}}/T_p)(\mathrm{ehr}/4)$, where $T_p$ is the number of image patch tokens
($196$ for S/16 and $256$ for B/14). For weight merging, $U{=}\mathrm{ehr}/4$.
We evaluate one merged FFN with $\mathrm{ehr}{=}4$, so the weight-merge methods remain at $U{=}1$.

\paragraph{Expert parameter storage:}
At a fixed $U$, methods may store different numbers of experts at different widths.
Table~\ref{tab:moe-storage} therefore reports the expert-FFN parameter counts at the S and B model
widths used in the supervised and distilled comparisons. For model width $d$ and expert hidden ratio
$r$, one expert contains $P_{\mathrm{FFN}}(d,r)=2rd^2+(r+1)d$ parameters, including both projections
and biases. We sum this quantity over all stored experts, including always-active shared experts. The
counts exclude the recurrent backbone, routers, and Soft-MoE slot embeddings. The weight-merging
methods store $E{=}4$ experts. Soft-MoE uses $49/64$ slots per expert for S/16 and B/14, respectively,
and $E{=}4,6,8,16$ experts at $U{=}1,1.5,2,4$.

\begin{table}[!ht]
\centering
\caption{\textbf{Stored expert-FFN parameters (millions).} Each entry gives the S/B counts.}
\label{tab:moe-storage}
\footnotesize
\setlength{\tabcolsep}{4pt}
\renewcommand{\arraystretch}{1.05}
\begin{tabular}{@{}lrrrr@{}}
\toprule
Method & $U{=}1$ & $U{=}1.5$ & $U{=}2$ & $U{=}4$ \\
\midrule
\multicolumn{5}{@{}l}{\textit{Weight merge}} \\
\methodname{} (ours) & 4.7/18.9 & -- & -- & -- \\
Lory                  & 4.7/18.9 & -- & -- & -- \\
SMEAR                  & 4.7/18.9 & -- & -- & -- \\
\midrule
\multicolumn{5}{@{}l}{\textit{Token dispatch}} \\
DeepSeek-V3            & 3.0/11.8 & 5.3/21.3 & 5.3/21.3 & 10.6/42.5 \\
Qwen3                  & 4.7/18.9 & 14.2/56.7 & 18.9/75.6 & 37.8/151.2 \\
MoEUT                  & 4.7/18.9 & 7.1/28.3 & 9.5/37.8 & 9.5/37.8 \\
\midrule
\multicolumn{5}{@{}l}{\textit{Output mixture}} \\
ReMoE                  & 1.2/4.7 & 1.8/7.1 & 2.4/9.4 & 4.7/18.9 \\
Soft-MoE               & 4.7/18.9 & 7.1/28.3 & 9.5/37.8 & 18.9/75.6 \\
\bottomrule
\end{tabular}
\end{table}

\begin{table}[!t]
\centering
\caption{\textbf{Complete distilled recurrent-MoE comparison.} Full results underlying
Table~\ref{tab:moe-dino}. All methods use the same B/14 recurrent backbone, $L{=}12$, and task loss. $U$
counts nominal expert-FFN arithmetic relative to one standard dense FFN per step. These fixed-depth
results are separate from the elastic-depth evaluation in \figref{fig:anytime-depth}. Higher-$U$ settings also change
the expert count, activation count, and/or width, as detailed below.}
\label{tab:moe-dino-full}
\footnotesize
\setlength{\tabcolsep}{4pt}
\renewcommand{\arraystretch}{1.05}
\begin{tabular}{@{}l cccccccc@{}}
\toprule
& \multicolumn{3}{c}{Weight merge} & \multicolumn{3}{c}{Token dispatch} & \multicolumn{2}{c@{}}{Output mixture} \\
\cmidrule(lr){2-4}\cmidrule(lr){5-7}\cmidrule(l){8-9}
$U$ & \rotatebox{90}{\shortstack[l]{\methodname{}~($E{=}4$)}}
 & \rotatebox{90}{\shortstack[l]{Lory\\[1pt]{\scriptsize\citep{zhong2024lory}}}}
 & \rotatebox{90}{\shortstack[l]{SMEAR\\[1pt]{\scriptsize\citep{muqeeth2023soft}}}}
 & \rotatebox{90}{\shortstack[l]{DeepSeek-V3\\[1pt]{\scriptsize\citep{liu2024deepseek}}}}
 & \rotatebox{90}{\shortstack[l]{Qwen3\\[1pt]{\scriptsize\citep{qwen2025qwen3}}}}
 & \rotatebox{90}{\shortstack[l]{MoEUT\\[1pt]{\scriptsize\citep{csordas2024moeut}}}}
 & \rotatebox{90}{\shortstack[l]{ReMoE\\[1pt]{\scriptsize\citep{wang2025remoe}}}}
 & \rotatebox{90}{\shortstack[l]{Soft-MoE\\[1pt]{\scriptsize\citep{puigcerver2024from}}}} \\
\midrule
\multicolumn{9}{@{}l}{\textit{ImageNet-1k linear top-1} (\%, $\uparrow$)} \\
\midrule
$1\times$   & \textbf{83.5} & 79.2 & 83.4 & 74.4 & -- & 75.1 & 79.6 & 78.5 \\
$1.5\times$ & -- & -- & -- & 71.4 & 71.5 & 76.1 & \textbf{81.0} & 79.9 \\
$2\times$   & -- & -- & -- & 75.9 & 73.6 & 75.2 & \textbf{81.5} & 80.7 \\
$4\times$   & -- & -- & -- & 76.6 & 74.9 & 83.0 & \textbf{83.2} & 82.2 \\
\midrule
\multicolumn{9}{@{}l}{\textit{ADE20k mIoU} (\%, $\uparrow$)} \\
\midrule
$1\times$   & \textbf{44.6} & 39.9 & 44.3 & 36.6 & -- & 35.0 & 39.7 & 33.4 \\
$1.5\times$ & -- & -- & -- & 39.3 & 38.9 & 40.3 & \textbf{41.1} & 34.4 \\
$2\times$   & -- & -- & -- & 38.1 & 38.9 & 41.3 & \textbf{42.1} & 36.3 \\
$4\times$   & -- & -- & -- & 39.9 & 40.1 & 43.5 & \textbf{45.1} & 37.1 \\
\midrule
\multicolumn{9}{@{}l}{\textit{NYUv2 RMSE, linear} (m, $\downarrow$)} \\
\midrule
$1\times$   & \textbf{0.552} & 0.580 & 0.553 & 0.654 & -- & 0.694 & 0.573 & 0.602 \\
$1.5\times$ & -- & -- & -- & 0.607 & 0.669 & 0.603 & \textbf{0.556} & 0.586 \\
$2\times$   & -- & -- & -- & 0.651 & 0.679 & 0.746 & \textbf{0.557} & 0.583 \\
$4\times$   & -- & -- & -- & 0.607 & 0.638 & 0.558 & \textbf{0.539} & 0.547 \\
\midrule
\multicolumn{9}{@{}l}{\textit{NYUv2 RMSE, MLP} (m, $\downarrow$)} \\
\midrule
$1\times$   & 0.491 & 0.531 & \textbf{0.485} & 0.582 & -- & 0.608 & 0.527 & 0.557 \\
$1.5\times$ & -- & -- & -- & 0.545 & 0.593 & 0.545 & \textbf{0.509} & 0.541 \\
$2\times$   & -- & -- & -- & 0.567 & 0.585 & 0.610 & \textbf{0.503} & 0.538 \\
$4\times$   & -- & -- & -- & 0.539 & 0.550 & 0.501 & \textbf{0.482} & 0.500 \\
\bottomrule
\end{tabular}
\end{table}

\paragraph{Lory:}
Lory uses a linear softmax router to merge expert parameters. In the original model, the pooled
representation of one token segment selects the FFN for the next segment, making routing causal
\citep{zhong2024lory}. A ViT processes its image tokens in parallel and has no equivalent token segmentation.
We therefore route across recurrent depth, using the pooled state from step $t{-}1$ to select the merged FFN
at step $t$. Apart from this depth reinterpretation, the routing and merge are unchanged. We omit
similarity-based document batching because it has no document-level analogue for individual images.
Since the merge produces one FFN, increasing $E$ changes stored capacity and adds parameter-synthesis
arithmetic equal to an $O(E/T)$ fraction of the dense FFN cost, but does not change $U$. Like
\methodname{}, Lory is evaluated only at $U{=}1$.

\paragraph{SMEAR:}
SMEAR merges adapter experts inside a frozen pretrained backbone, takes one routing decision per example,
and deliberately omits load-balancing losses \citep{muqeeth2023soft}.
In our controlled comparison, the SMEAR row applies a content-conditioned softmax mixture to standard FFNs and uses no auxiliary losses. It therefore differs from \methodname{} in its conditioning signal, not in the broader parameter-composition family. Table~\ref{tab:moe-dino} reports the fixed-depth result. The corresponding feature-only control in \figref{fig:anytime-depth} uses the same depth-dropout protocol as our depth-only variant. Its feature-conditioned gate is recomputed at every recurrent depth. At $E{=}4$ and $197$ tokens, the merge costs about $E/T\approx2\%$ of
the arithmetic required to apply the FFN to all tokens. Because adding experts does not change $U$, the only way to increase $U$ in this adaptation is to widen the merged FFN.

\paragraph{DeepSeek-V3:}
We retain three components of DeepSeek-V3's routing design: sigmoid affinities, one always-active shared expert,
and the auxiliary-loss-free bias balancer, which adds a per-expert bias to the affinity before top-$K$
selection and never to the gate value \citep{liu2024deepseek,wang2024auxiliary}. At $U{=}1$, our recurrent DeepSeek-V3 variant uses four routed experts with one active per token, plus one always-active shared expert, all at $\mathrm{ehr}{=}2$. DeepSeek-V3 instead activates 8 of 256 much narrower routed experts at an expert hidden ratio near $0.29$. For
$U{>}1$, this variant jointly increases the routed- and shared-expert counts: $E{=}8$ with $K{=}2$ plus $1$ shared at
$\mathrm{ehr}{=}2$, then $E{=}16$ with $K{=}6$ plus $2$ shared at $\mathrm{ehr}{=}1$, then $E{=}32$ with $K{=}12$ plus $4$ shared at
$\mathrm{ehr}{=}1$. The $U{=}2$ and $U{=}4$ configurations keep the shared-to-routed ratio close to DeepSeekMoE's $1{:}3$ \citep{dai2024deepseekmoe}, while the $U{=}1.5$ configuration uses $1{:}2$. DeepSeek-V3 instead keeps one shared expert at every scale. We omit V3's sequence-level balance loss. We also omit its leading dense layers because a fully weight-tied encoder applies the same MoE block at all $12$ steps and therefore cannot reserve only the early layers for dense computation.

\paragraph{Qwen3:}
Our adaptation retains Qwen3's routing rule: softmax over all experts, top-$K$ selection with renormalization, no shared expert, and the global-batch load-balancing loss with coefficient $10^{-3}$ in the higher-budget configurations \citep{qwen2025qwen3,qiu2025demons}. At $U{=}1$, it activates one of four experts with $\mathrm{ehr}{=}4$. Under distillation, this configuration failed to converge to a usable solution, so we report no downstream results for it. Higher-budget configurations use $\mathrm{ehr}{=}1$ and keep $12.5\%$ of experts active: 6 of 48, 8 of 64, and 16 of 128. Qwen3 itself activates 8 of 128 experts ($6.25\%$).

\paragraph{MoEUT:}
MoEUT typically activates about $16$ narrow experts from a bank of hundreds and sums their outputs to
approximate one wide FFN \citep{csordas2024moeut}. Our $U{=}1$ adaptation instead activates one of four
standard FFN experts. Higher-budget variants activate $2$ of $8$ experts at $\mathrm{ehr}{=}3$, $4$ of
$16$ at $\mathrm{ehr}{=}2$, and $16$ of $32$ at $\mathrm{ehr}{=}1$. Only the $U{=}4$ configuration reaches
the $K{=}16$ regime used by MoEUT. We retain dense attention and omit SwitchHead and peri-LayerNorm.

\paragraph{ReMoE:}
We retain ReMoE's ReLU gates and sparsity controller, using target sparsity $1-K/E{=}0.75$,
$\lambda_0{=}10^{-8}$, and $\alpha{=}1.2$ \citep{wang2025remoe}. ReMoE uses dense FFN experts, while its
fine-grained variant preserves active capacity by increasing the total and active expert counts from
$E,K$ to $EG,KG$. Our recurrent adaptation instead uses four experts with $\mathrm{ehr}{=}1$, evaluates
all four, and combines their outputs using ReMoE's gates. This retains its routing rule but turns the layer
into a dense output mixture. At higher budgets, $\mathrm{ehr}$ remains $1$, while $E$ increases to $6$,
$8$, and $16$. Because every expert is evaluated, $U$ grows directly with $E$.

\paragraph{Soft-MoE:}
Our adaptation retains full-width slot experts and the $\ell_2$ normalization of router input and router weights
\citep{puigcerver2024from}. At $U{=}1$, S/16 uses $196$ slots and B/14 uses $256$, split across
$E{=}4$ experts with $49$ and $64$ slots per expert, respectively.
The original Soft-MoE ablation favors one slot per expert, but with four stored experts that setting would
fall well below the matched $U{=}1$ budget. At $U{=}1.5,2,4$, S/16 uses $294$, $392$, and $784$ slots,
while B/14 uses $384$, $512$, and $1024$. These exceed the respective patch counts and increase nominal
computation while the input token count remains fixed. The original work also studies increasing the number of slots. The original model applies Soft-MoE
only to the second half of its MLP blocks, whereas full
weight tying makes every recurrent step a Soft-MoE step.

\subsection{Layer-wise LoRA baseline}
\label{app:lora-baseline}

\begin{table}[!ht]
\centering
\caption{\textbf{Depth-specific parameterizations on ImageNet-1k.} Parameter counts refer to compact
checkpoints. Online GFLOPs include parameter construction for \methodname{} and low-rank updates for LoRA.
Folded GFLOPs use $12$ precomputed depth-specific dense weight sets. The \methodname{} result is the mean
of three runs. LoRA and DeiT~III are single runs trained with the same supervised recipe.}
\label{tab:lora-baseline}
\footnotesize
\setlength{\tabcolsep}{2pt}
\begin{tabular}{@{}lrrrc@{}}
\toprule
Method & \shortstack{Compact\\params (M)} & \multicolumn{1}{c}{GFLOPs} & Top-1 (\%) $\uparrow$ & Depth index \\
       &            & \multicolumn{1}{c}{online / folded} & & \\
\midrule
DeiT~III-S/16 & 22.1 & 8.5 / 8.5 & 79.9 & Block index \\
Tied ViT-S/16 + LoRA, $r{=}64$ & 7.3 & 10.4 / 8.5 & 77.6 & Adapter index \\
\methodname{}-S/16, $E{=}4$ & 6.1 & 8.6 / 8.5 & 78.2 & Coordinate \\
\bottomrule
\end{tabular}
\end{table}

In Table~\ref{tab:lora-baseline}, we compare \methodname{} with a recurrent ViT-S/16 using the layer-wise LoRA parameterization of
\citet{bae2025relaxed}. The model reuses one block for $L{=}12$ depths and adds an independent rank-$64$
update $\mW_t=\mW+\mB_t\mA_t$ to the fused QKV projection, attention output projection, and both FFN
projections at each depth. The base weights, biases, and normalization parameters remain shared. We train
the shared block and adapters jointly from random initialization using the $300$-epoch supervised S/16
recipe in Table~\ref{tab:hyperparameters}.

Layer-wise LoRA reaches $77.6\%$ top-1, compared with $78.2\%$ for \methodname{}, while storing $7.3$M
rather than $6.1$M compact parameters. At width $d{=}384$, its $12$ adapter sets add
$16\times Lrd=4.72$M parameters to the $2.53$M-parameter shared model. Each adapter corresponds to one trained
depth, so changing the number of recurrent steps requires a rule for selecting or interpolating adapters.
\methodname{} instead evaluates its continuous normalized-depth program on the new grid, as tested in
\figref{fig:anytime-depth}.

\section{Regularization ablations}
\label{app:training-ablations}

\paragraph{Router z-loss and balance loss:}
Table~\ref{tab:router-losses} varies the z-loss and usage-balance coefficients one at a time while holding
the other at its default. Accuracy remains within $0.6$ ImageNet-1k top-1 points of the default across all
tested values, with the default giving the highest observed accuracy in both sweeps.

\begin{center}
\small
\captionsetup{hypcap=false}
\captionof{table}{\textbf{Router-regularization sensitivity.} Each panel varies one coefficient while the
other remains at its default. Entries are changes in ImageNet-1k top-1 (percentage points) from
$(\lambda_z,\lambda_{\mathrm{bal}}){=}(10^{-3},10^{-2})$.}
\label{tab:router-losses}
\begin{minipage}[t]{0.46\textwidth}
\centering
{\bf (a) Router z-loss $\lambda_z$}\\[3pt]
\begin{tabular}{lc}
\toprule
$\lambda_z$ & $\Delta$ top-1 \\
\midrule
$0$                 & $-0.2$ \\
$10^{-4}$           & $-0.3$ \\
$10^{-3}$ (default) & $\phantom{-}0.0$ \\
$10^{-2}$           & $-0.4$ \\
\bottomrule
\end{tabular}
\end{minipage}
\hfill
\begin{minipage}[t]{0.46\textwidth}
\centering
{\bf (b) Usage-balance loss $\lambda_{\mathrm{bal}}$}\\[3pt]
\begin{tabular}{lc}
\toprule
$\lambda_{\mathrm{bal}}$ & $\Delta$ top-1 \\
\midrule
$0$                 & $-0.4$ \\
$10^{-3}$           & $-0.3$ \\
$10^{-2}$ (default) & $\phantom{-}0.0$ \\
$10^{-1}$           & $-0.6$ \\
\bottomrule
\end{tabular}
\end{minipage}
\end{center}

\section{Additional representation and expert analyses}
\label{app:analysis}

\begin{figure}[!ht]
  \centering
  \includegraphics[width=0.95\linewidth]{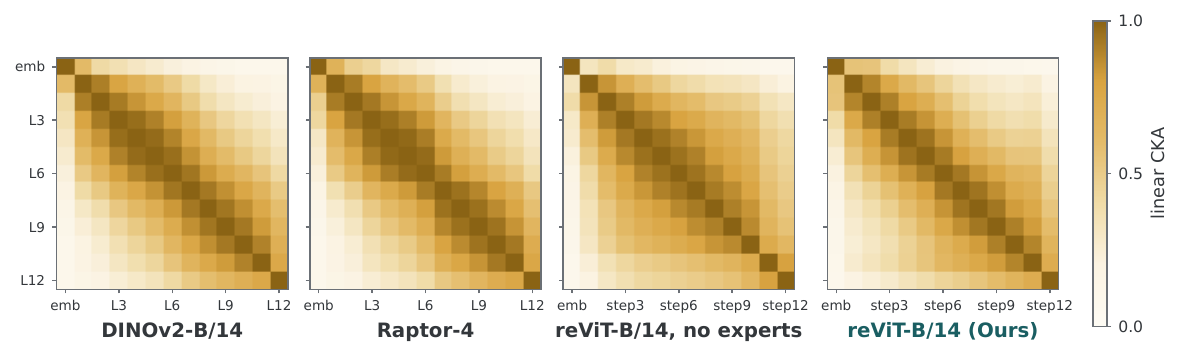}
  \caption{\textbf{Depthwise self-CKA after DINOv2 distillation.} Linear CKA of mean-pooled patch-token
  residual states from the embedding through depth 12, computed over 1,000 ImageNet validation images,
  for DINOv2-B/14, Raptor-$4$, and distilled \methodname-B/14 with $E\in\{1,8\}$ and $L{=}12$.
  The $E{=}1$ model is a plain tied block without an expert bank.}
  \label{fig:cka}
\end{figure}

\begin{figure}[!h]
  \centering
  \includegraphics[width=\linewidth]{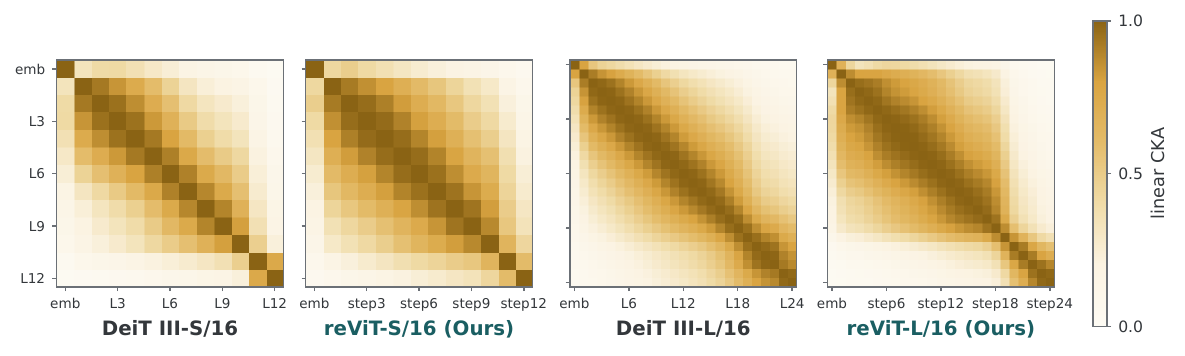}
  \caption{\textbf{Depthwise self-CKA after supervised ImageNet-1k training.} Linear CKA of mean-pooled
  patch-token residual states over 1,000 validation images for DeiT~III-S/16 and \methodname-S/16
  (12 blocks/steps), and DeiT~III-L/16 and \methodname-L/16 (24 blocks/steps). Dark near-diagonal bands
  indicate greater similarity between neighboring depths.}
  \label{fig:cka-sup}
\end{figure}

\begin{figure}[t]
  \centering
  \includegraphics[width=\linewidth]{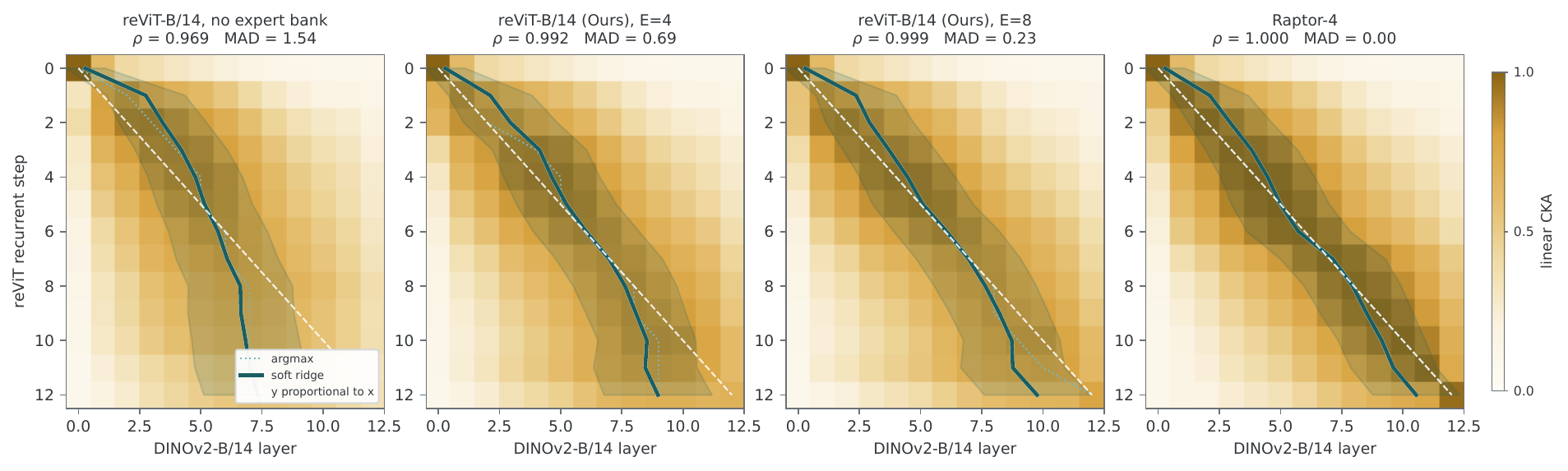}
  \caption{\textbf{Cross-CKA with DINOv2 across depth.} Linear CKA between DINOv2-B/14 states (horizontal)
  and student states (vertical) for distilled \methodname-B/14 with $E\in\{1,4,8\}$ and Raptor-$4$, over the
  same 1,000-image subset as Figs.~\ref{fig:cka} and~\ref{fig:cka-sup}. Dotted lines mark the maximum-CKA
  teacher depth, and dashed lines mark proportional depth. $\rho$ is the Spearman correlation between
  recurrent depth and the maximum-CKA teacher depth. MAD is that path's mean absolute deviation from
  proportional depth in teacher-layer units. Solid lines and bands show the weighted mean $\pm$ one
  standard deviation using weights proportional to $\max(\mathrm{CKA},0)^4$.}
  \label{fig:ridge}
\end{figure}

\begin{figure}[t]
  \centering
  \begin{minipage}[t]{0.495\linewidth}
    \centering
    \includegraphics[width=\linewidth]{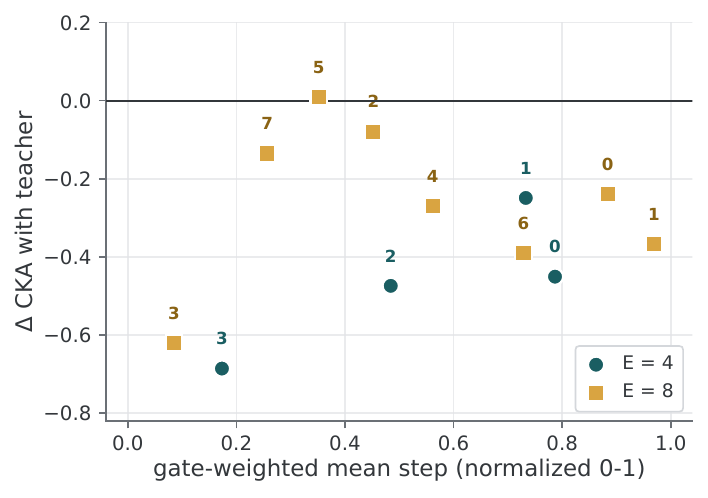}\\[1pt]
    {\bf (a)} Ablation effect versus mean depth
  \end{minipage}
  \hfill
  \begin{minipage}[t]{0.495\linewidth}
    \centering
    \includegraphics[width=\linewidth]{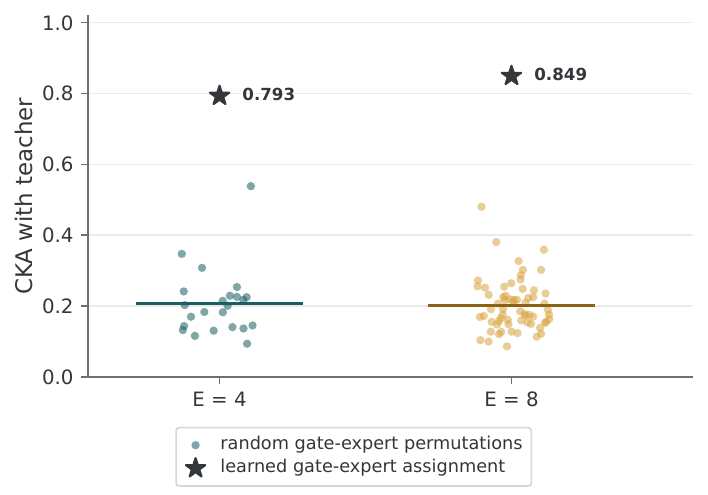}\\[1pt]
    {\bf (b)} Gate--expert assignment permutation
  \end{minipage}
  \vspace{-6bp}
  \caption{\textbf{Schedule controls for expert ablation.} (a) CKA change after ablation against each
  expert's gate-weighted mean normalized depth. (b) Teacher CKA for the learned assignment (stars) and
  nonidentity permutations of gate columns across the fixed expert bank (dots). Horizontal lines show
  permutation means. We evaluate all 23 permutations for $E{=}4$ and 64 sampled permutations for $E{=}8$.}
  \label{fig:drop-controls}
\end{figure}

\paragraph{Common evaluation protocol:}
The self- and cross-CKA analyses in Figs.~\ref{fig:cka}, \ref{fig:cka-sup}, and~\ref{fig:ridge} use a
fixed subset of $1{,}000$ ImageNet validation images selected with a fixed seed, bicubically resized to
$256$ and center-cropped to $224$. The ablation and permutation analyses in
Figs.~\ref{fig:drop}, \ref{fig:drop-controls}
use $256$ images from the same validation set. For every model, we record the residual stream after the
embedding and after each block or recurrent depth, before the final LayerNorm. We remove prefix tokens,
average the patch-token representations, and compare the resulting features using linear CKA
\citep{kornblith2019similarity}. The routing and rank analyses are computed directly from the learned
router and expert weights and do not depend on evaluation images.

\paragraph{Effective-rank protocol:}
For \figref{fig:rank}, let $P=d_{\mathrm{ff}}d$. We form
$\mW_{1,\mathrm{bank}}\in\R^{E\times P}$ by stacking the vectorized first-projection matrices, with row
$e$ equal to $\operatorname{vec}(\mW_1^e)^\top$. The gate matrix $\mG\in\R^{L\times E}$ has entries
$G_{t,e}=g_{t,e}$. Stacking the corresponding merged projections across depth gives
$\mW_{1,\mathrm{traj}}=\mG\mW_{1,\mathrm{bank}}$. Therefore,
\[
\operatorname{rank}(\mW_{1,\mathrm{traj}})
\leq
\min\{\operatorname{rank}(\mG),
      \operatorname{rank}(\mW_{1,\mathrm{bank}})\}
\leq E.
\]
For each matrix, with singular values $\sigma_i$, we use $p_i=\sigma_i^2/\sum_j\sigma_j^2$ and
$r_{\mathrm{eff}}=\exp(-\sum_i p_i\log p_i)$ \citep{roy2007effective}.

To remove the component shared across depth, we define $\mC=\mI_L-L^{-1}\1\1^\top$ and compute
$\mG_c=\mC\mG$ and
$\mW_{1,\mathrm{traj},c}=\mC\mW_{1,\mathrm{traj}}=\mG_c\mW_{1,\mathrm{bank}}$. Because every row of
$\mG$ sums to one, both centered matrices have rank at most $E-1$.

\begin{figure}[!ht]
  \centering
  \begin{minipage}[t]{0.495\linewidth}
    \centering
    \includegraphics[width=\linewidth]{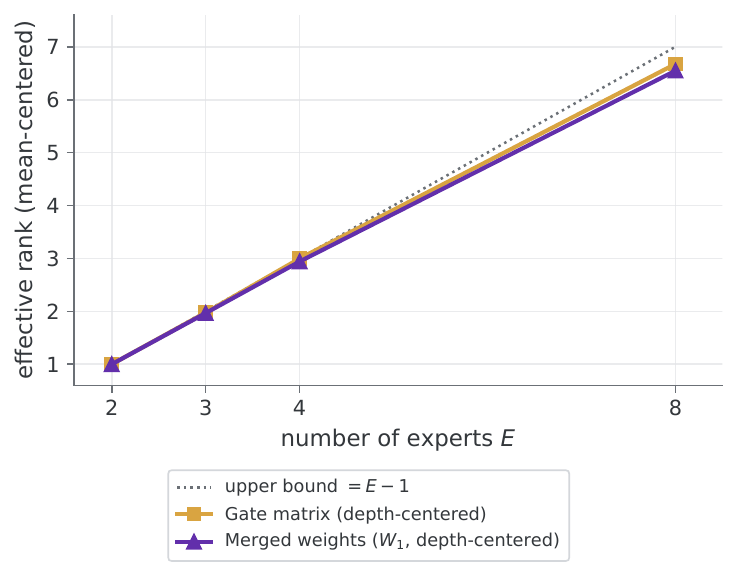}\\[1pt]
    {\bf (a)} \methodname-S/16 ($L{=}12$)
  \end{minipage}
  \hfill
  \begin{minipage}[t]{0.495\linewidth}
    \centering
    \includegraphics[width=\linewidth]{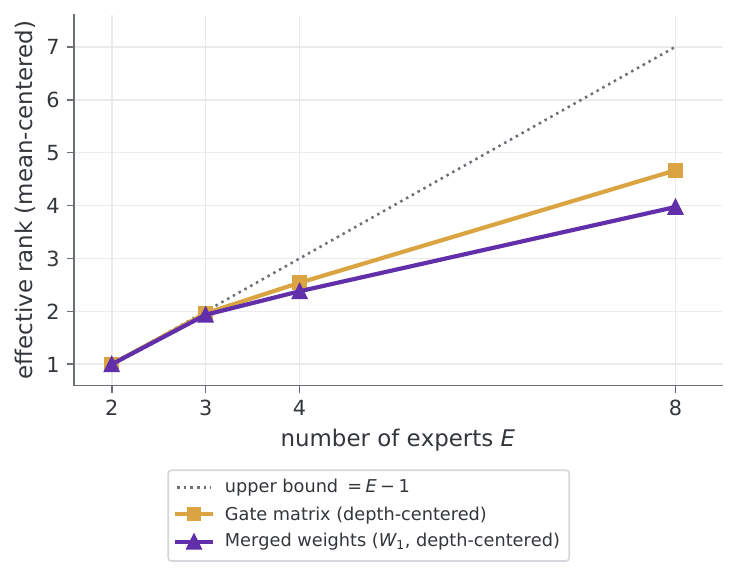}\\[1pt]
    {\bf (b)} \methodname-L/16 ($L{=}24$)
  \end{minipage}
  \vspace{-6bp}
  \caption{\textbf{Depth-centered routing and merged-weight ranks.}
  Centering removes the mean gate and merged $W_1$ vectors across recurrent depth. The dashed line marks
  the resulting rank ceiling $E-1$.}
  \label{fig:rank-centered}
\end{figure}

\paragraph{Banded self-similarity is a property of recurrence:}
Self-CKA describes the trajectory within each model (\figref{fig:cka}). DINOv2-B/14 has high similarity
between neighboring layers that declines with depth separation, as previously observed in ViTs
\citep{raghu2021vision}. The same broad pattern appears in Raptor and both recurrent \methodname{} models.
In particular, the plain $E{=}1$ tied block remains banded despite having no expert bank. Recurrence alone
therefore produces a noncollapsed depthwise trajectory, and this pattern should not be attributed to
expert routing.

\FloatBarrier

\paragraph{The banded structure also emerges under supervised training:}
\Figref{fig:cka-sup} extends the analysis of \figref{fig:cka} to the supervised regime, comparing our
S/16 and L/16 models with DeiT~III baselines \citep{touvron2022deit3} of the same size. The same banded
structure appears at both scales. For L/16, the band of our model narrows visibly over the last quarter
of the recurrence. We treat these self-CKA panels as descriptions of the recurrent trajectories rather
than evidence for the expert mechanism.

Self-CKA describes the structure within each trajectory. We next use cross-model CKA to compare student
steps with teacher layers.

\FloatBarrier

\paragraph{Expert capacity improves alignment to the teacher's depth progression:}
Within the recipe-matched \methodname{} family, the maximum-CKA path approaches proportional teacher
depth as the expert bank grows. Its $\rho$/MAD improves from $0.969/1.54$ at $E{=}1$ to $0.992/0.69$ at
$E{=}4$ and $0.999/0.23$ at $E{=}8$ (\figref{fig:ridge}). The $E{=}1$ path saturates at teacher layer 7,
whereas $E{=}8$ reaches layer 12. Excluding the embedding and final states preserves the ordering, with
interior-step MAD values of $1.36$, $0.55$, and $0.27$. Thus, additional expert capacity improves how
evenly the recurrent trajectory covers teacher depth despite supervision only from the teacher's final
output features.

Raptor's maximum-CKA path follows the proportional diagonal exactly. This is expected because its
training objective directly supervises intermediate teacher layers \citep{jacobs2026block}. We therefore treat it as a reference
for explicit intermediate alignment rather than a control for alignment emerging under output-only
distillation.

\paragraph{Expert ablations are shaped by schedule position:}
We ablate one expert by setting its gate weight to zero at every depth, renormalizing the remaining
weights, and leaving the checkpoint fixed. We report $\Delta\mathrm{CKA}=\mathrm{CKA}_{\mathrm{ablation}}-
\mathrm{CKA}_{\mathrm{intact}}$, so negative values indicate worse teacher alignment. For each ablation, we draw 64 random gate perturbations
matched to its Frobenius displacement in the merged FFN weights. Most expert ablations are less damaging
than random changes of the same size (\figref{fig:drop}). At $E{=}8$, removing expert 5 changes
CKA by $+0.009$, with no measurable adverse effect.

Raw ablation cost also depends on when an expert is used. The earliest-routed experts cause the largest
losses (\figref{fig:drop-controls}a).

\FloatBarrier

Ablation magnitude alone therefore does not establish an expert-specific role. We test the assignment
directly by permuting complete gate columns across the fixed expert bank. This retains every expert and
gate trajectory while changing only their pairing. Teacher CKA falls from $0.793$ to $0.208$ on average
for $E{=}4$ and from $0.849$ to $0.202$ for $E{=}8$, and no tested permutation reaches the learned
assignment (\figref{fig:drop-controls}b).

\clearpage
\bibliographystyle{plainnat}
\bibliography{references}

\end{document}